\documentclass[10pt, journal]{IEEEtran}
\usepackage{amsmath,amsfonts}
\usepackage{algorithmic}
\usepackage{algorithm}
\usepackage{array}
\usepackage[caption=false,font=scriptsize,labelfont=sf,textfont=sf]{subfig}
\usepackage{textcomp}
\usepackage{stfloats}
\usepackage{url}
\usepackage{verbatim}
\usepackage{graphicx}
\usepackage{cite}

\usepackage{amsmath,amsfonts}
\usepackage{algorithm}
\usepackage{array}
\usepackage{textcomp}
\usepackage{stfloats}
\usepackage{url}
\usepackage{verbatim}
\usepackage{graphicx}
\usepackage{multirow}
\usepackage{amsfonts}
\usepackage{amssymb}
\usepackage{amsthm,amsmath}
\usepackage{mathrsfs}
\usepackage{indentfirst}
\usepackage{cite}
\usepackage{makecell}
\usepackage{float}

\newtheorem{lemma}{Lemma}
\usepackage[normalem]{ulem}
\useunder{\uline}{\ul}{}
\usepackage{xcolor}

\begin{document}

\title{Semantics-enhanced Temporal Graph Networks for
Content Popularity Prediction}

\author{\IEEEauthorblockN{Jianhang Zhu, Rongpeng Li, Xianfu Chen, Shiwen Mao, Jianjun Wu and Zhifeng Zhao}

\thanks{
   Jianhang Zhu and Rongpeng Li are with the College of Information Science and Electronic Engineering, Zhejiang University, Hangzhou 310027, China (e-mail: \{zhujh20; lirongpeng\}@zju.edu.cn).

   Xianfu Chen is with VTT Technical Research Centre of Finland, 90570 Oulu, Finland (e-mail: xianfu.chen@vtt.fi).

   Shiwen Mao is with Department of Electrical and Computer Engineering, Auburn University, Auburn, AL 36849-5201, USA (email: smao@ieee.org).
   
   Jianjun Wu is with Huawei Technologies Company, Ltd., Shanghai 201206, China (e-mail: wujianjun@huawei.com).
   
   Zhifeng Zhao is with Zhejiang Lab, Hangzhou, China as well as the College of Information Science and Electronic Engineering, Zhejiang University, Hangzhou 310027, China (e-mail: zhaozf@zhejianglab.com).
  }
}

\maketitle

\begin{abstract}
  The surging demand for high-definition video streaming services and large neural network models (e.g., Generative Pre-trained Transformer, GPT) implies a tremendous explosion of Internet traffic.
  To mitigate the traffic pressure, architectures with in-network storage have been proposed to cache popular contents at devices in closer proximity to users.
  Correspondingly, in order to maximize caching utilization, it becomes essential to devise an effective popularity prediction method. 
  In that regard, predicting popularity with dynamic graph neural network (DGNN) models achieve remarkable performance.
  However, DGNN models still suffer from tackling sparse datasets where most users are inactive. 
  Therefore, we propose a reformative temporal graph network, named semantics-enhanced temporal graph network (STGN), which attaches extra semantic information into the user-content bipartite graph and could better leverage implicit relationships behind the superficial topology structure.
  On top of that, we customize its temporal and structural learning modules to further boost the prediction performance.
  Specifically, in order to efficiently aggregate the diversified semantics that a content might possess, we design a user-specific attention (\texttt{UsAttn}) mechanism for temporal learning module.
  Unlike the attention mechanism that only analyzes the influence of genres on content, \texttt{UsAttn} also considers the attraction of semantic information to a specific user.
  Meanwhile, as for the structural learning, we introduce the concept of positional encoding into our attention-based graph learning and adopt a semantic positional encoding (\texttt{SPE}) function to facilitate the analysis of content-oriented user-association analysis.
  Finally, extensive simulations verify the superiority of our STGN models and demonstrate the effectiveness in content caching.

\end{abstract}

\begin{IEEEkeywords}
  Content caching, popularity prediction, dynamic graph neural network, semantics.
\end{IEEEkeywords}
  
\section{Introduction}\label{sec1}
  \IEEEPARstart{T}{HE} surging demand for high-definition video streaming services and large neural network models results in a tremendous pressure on the Internet \cite{cisco2020cisco, meybodi2022tedge,9134426}. 
  It is pointed out that caching popular contents or models in advance has the potential to reduce the backhaul traffic up to 35\% \cite{ETSI}.
  Correspondingly, in-network caching emerges as a promising technique and garners extensive attention \cite{7194828, 9427180}. 
  However, compared with the continual explosion of content volume, it is infeasible to increase the device caching capability immoderately due to the practical limitations (e.g., economic and technical perspectives) \cite{9187344}. 
  This predicament makes the design of competent caching strategies much more crucial, wherein accurate popularity prediction plays a decisive role \cite{9187344}. 

  Recently, deep neural networks (DNNs) have demonstrated their remarkable potential by unveiling the embedded temporal correlation for popularity prediction \cite{8172025, 9234632, 9846902}. 
  Meanwhile, along with users requesting contents, the interactions between users and contents gradually constitute a dynamic bipartite interaction graph.
  Some recent graph neural network (GNN) model-based methods, which resort to exploiting the inherent structural pattern in such a bipartite graph, manifest themselves in providing superior prediction accuracy within a recommendation system (RS) \cite{zhou2020graph, wu2022graph}.
  In particular, such GNN models enable us to speculate for inactive users with few requests in an interaction-intense graph by associating them with other active users that exhibit similar behaviors. 
  However, these GNN models \cite{zhou2020graph, wu2022graph} are contingent on an assumption of a static bipartite graph.
  In order to blend the merits of both structural learning and temporal learning, recommendation with dynamic graph neural network (DGNN) models emerges \cite{skarding2021foundations}, which is always synergistic with caching \cite{9420319}.
  Thus, caching with DGNN models also achieves satisfactory improvement \cite{zhu2022aoi}.

  Nevertheless, Ref. \cite{zhu2022aoi} discovers that the model's performance is not gratifying for cases where most users in the bipartite graph are inactive.
  To overcome the data sparsity, enlarging the receptive field of the DGNN model (e.g., the stacking of GNN layers or an increase in the number of first-order neighbors) might be productive \cite{Xu2020Inductive}, but it incurs significant computational complexity and time cost as well.
  On the contrary, along with fine-grained methods, enlarging the breadth of available information, such as considering the side information of users and contents in DGNN, sounds more appealing \cite{9220908, liu2021contextualized}.
  Besides, the side information enriches the sparse graph and may also reflect user's intention, both of which would benefit the reasoning and interpretability of user's future behavior.
  But given the importance of user privacy, it is more appropriate to conduct an excavation for the general content information (e.g., analyzing the semantic correlations among the genre information of contents).
 
\begin{figure}[tbp]
    \centering 
    \includegraphics[scale =0.4]{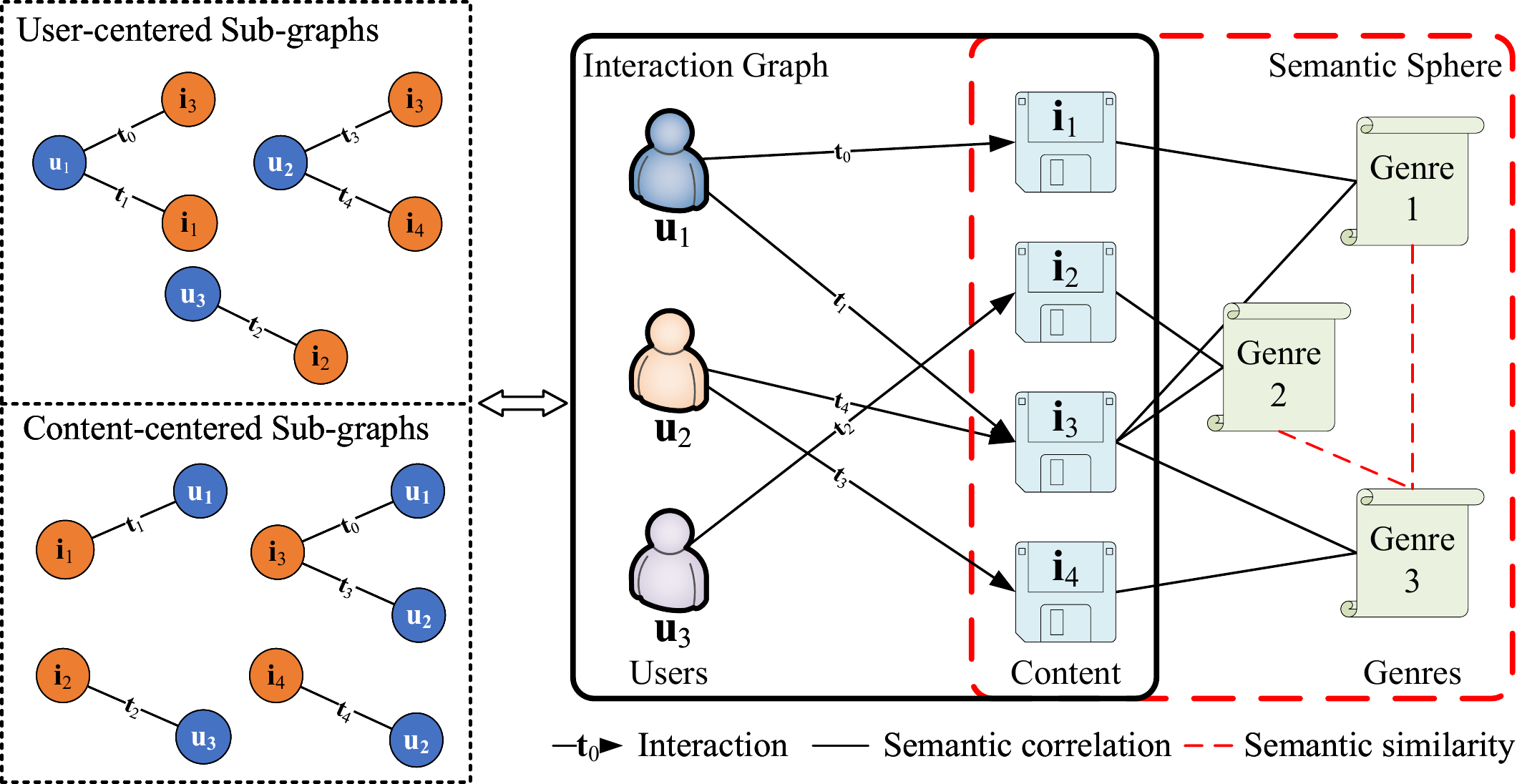}
    \caption{An example of the dynamic interaction graph and the implicit semantic relationship between contents.}
    \vspace{-1.2em}
    \label{DG}
 \end{figure}

\begin{figure*}[tbp]
  \centering
    \subfloat[One-hot]{\includegraphics[width = 0.325\textwidth]{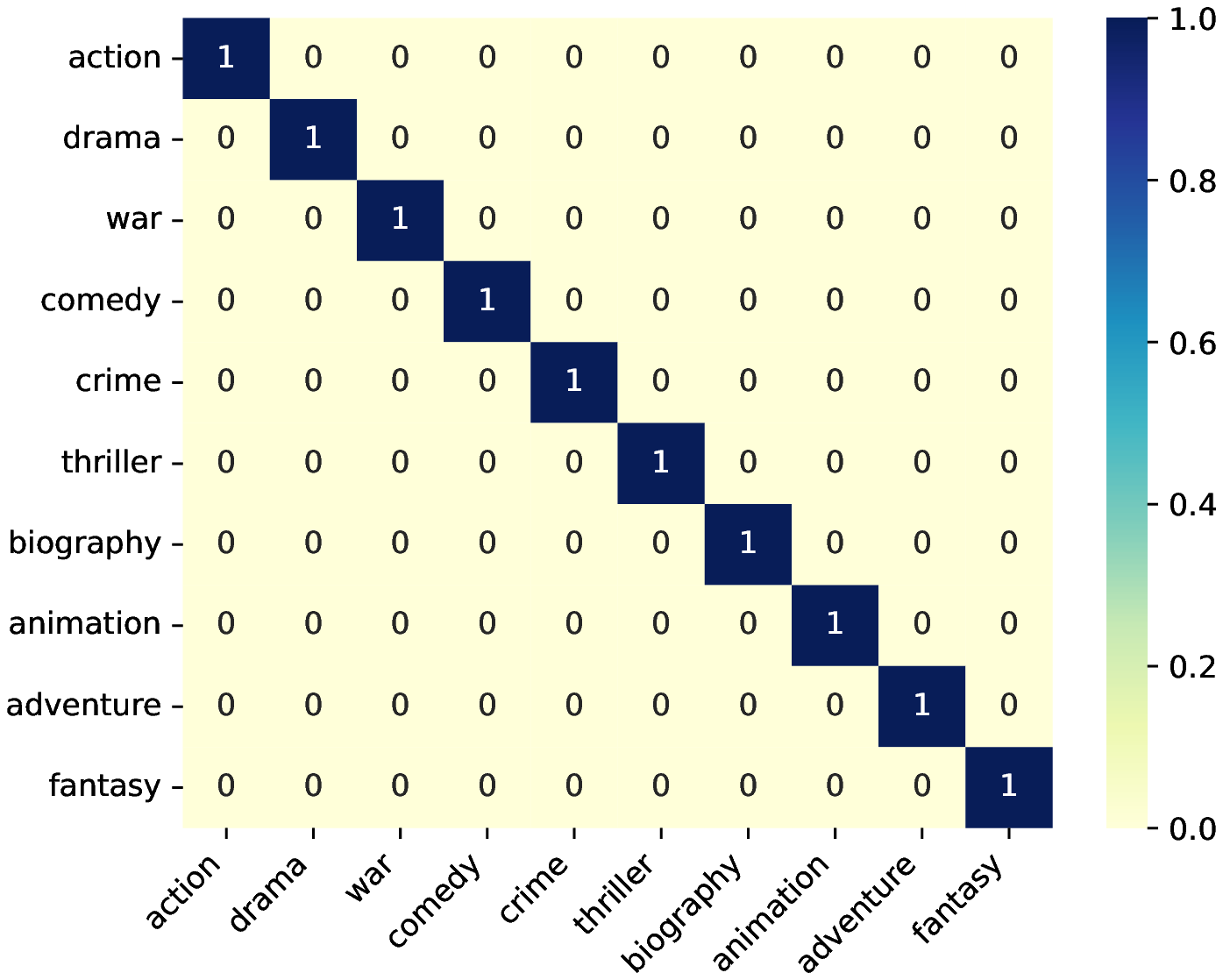}}
    \hspace{1mm} 
    \subfloat[BERT]{\includegraphics[width = 0.325\textwidth]{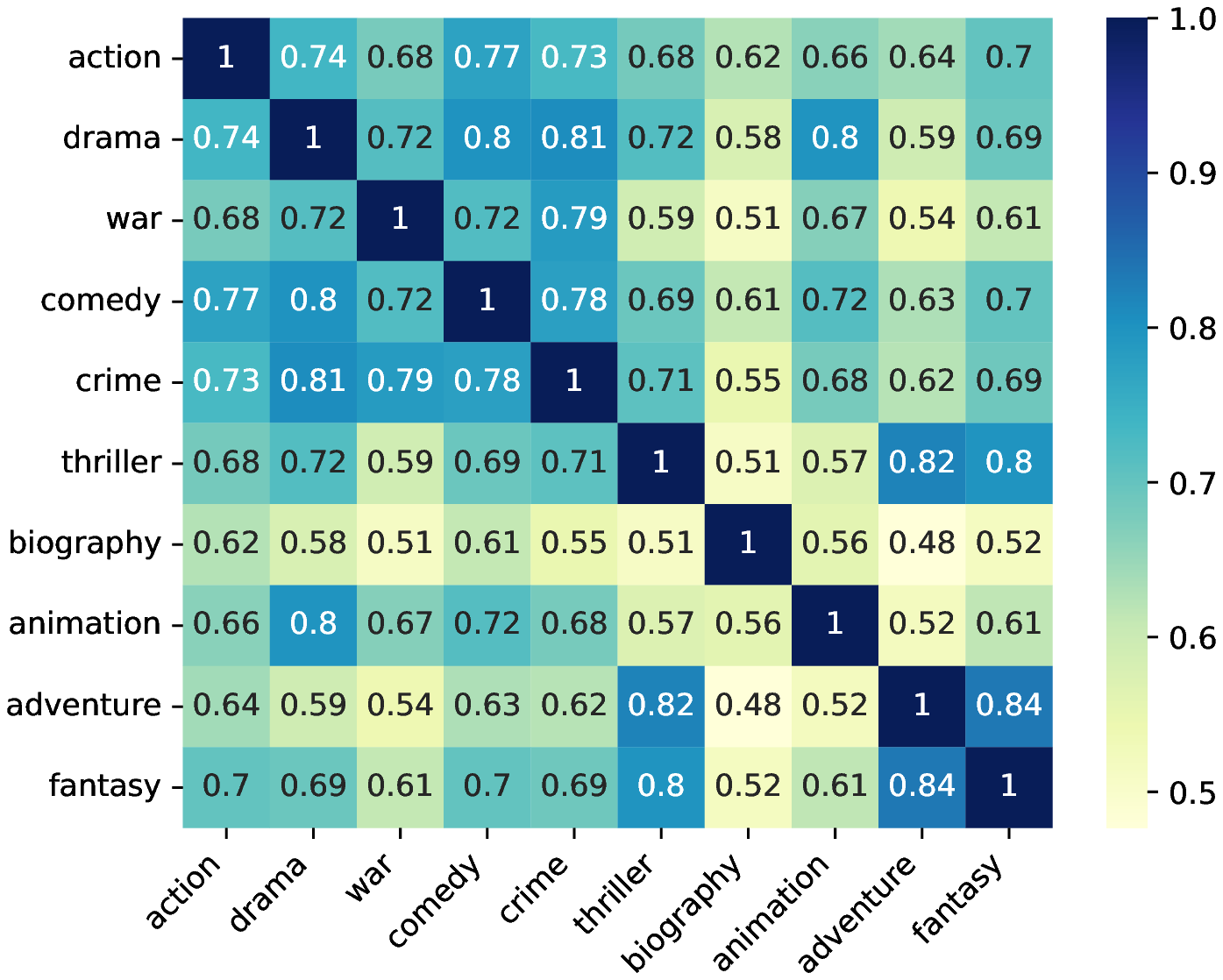}}
    \hspace{1mm} 
    \subfloat[Glove]{\includegraphics[width = 0.325\textwidth]{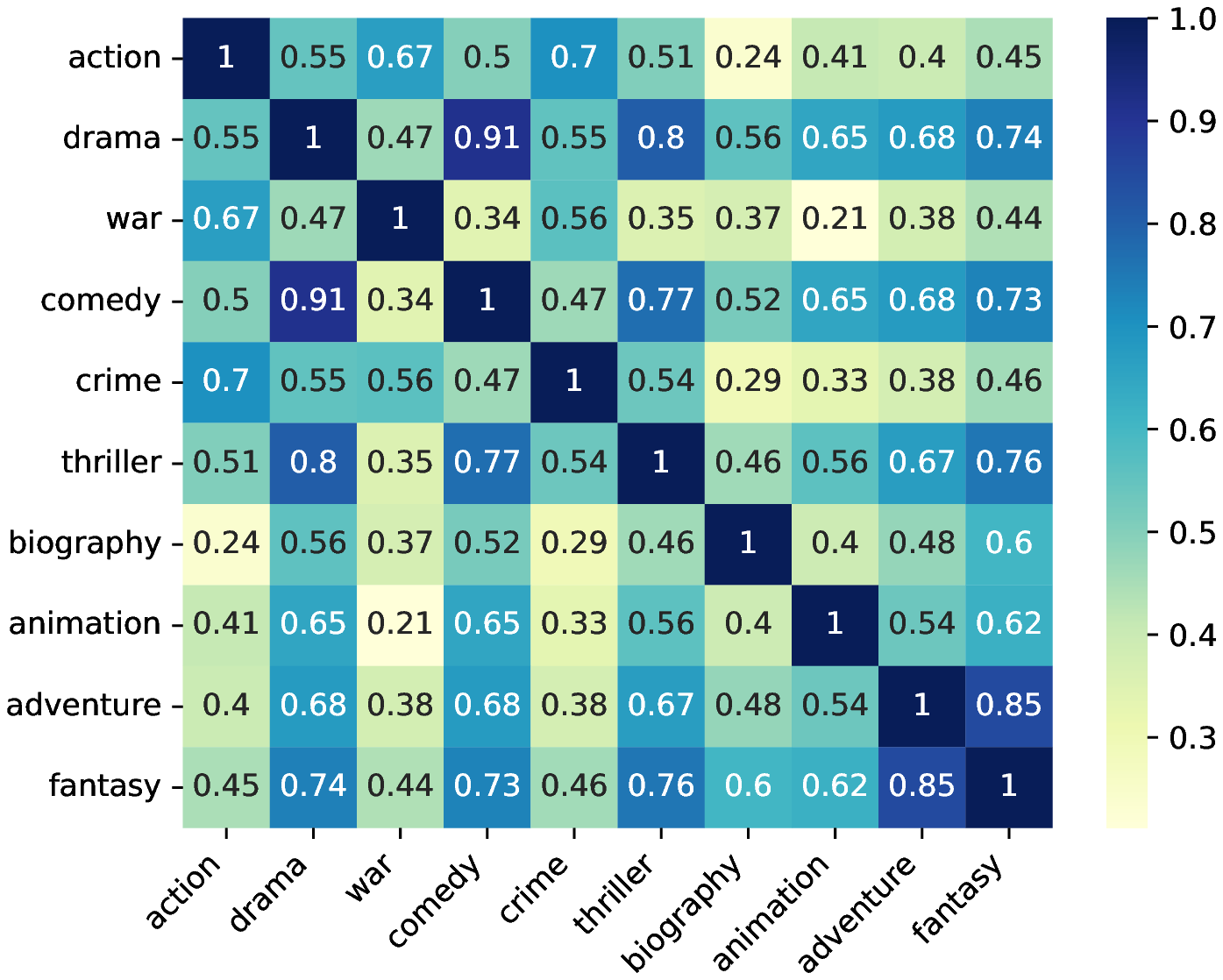}} 
  \caption{The visualization of cosine similarities between some genres' embeddings computed by different NLP methods.}
  \vspace{-1.2em}
  \label{fig:seman}
 \end{figure*}

  Fig. \ref{DG} presents an example of the dynamic interaction graph and the implicit semantic relationship between contents. 
  The requests of two users, $u_1$ and $u_2$, only intersect at content $i_3$. 
  The sparsity of data makes it intractable for classical GNN-based methods to accurately predict the preference of $u_1$ for content $i_4$. 
  On the other hand, the content genre information and their underlying similarities in the semantic sphere, which are indicated by the red dotted lines in Fig. \ref{DG}, reveal a strong correlation between $i_4$ and $i_3$ as well as a weak correlation between $i_4$ and $i_1$.
  Thanks to the attachment of semantics, these two kinds of underlying connectivity between $i_4$ and $i_1$ are unveiled to assist the prediction, and an inference that $u_1$ is likely to request $i_4$ can be boldly triggered.
  Furthermore, as demonstrated in Fig. \ref{fig:seman}, there exist several natural language processing (NLP) methods, such as one-hot, BERT \cite{DBLP:conf/naacl/DevlinCLT19}, and Glove \cite{pennington2014glove}, available for computing the semantic similarities.
  It is also natural to conjecture that a more precise computation of semantic similarities benefit to a superior speculation.

  In this paper, we propose a semantics-enhanced temporal graph network (STGN) to strengthen the DGNN model performance in dealing with sparse datasets and realize the semantics-assisted popularity prediction in content caching task.
  In particular, semantics ameliorates the temporal learning module to better track the dynamical variations in a user-content bipartite graph, and circumvents the difficulties in discovering patterns in the sparse dataset attributed to the supplement to content-centered sub-graphs for the structural learning. 
  Besides, we adopt some mature NLP methods to encode genre messages as semantic information, and then treat the embedded semantic information as part of the input of the predictive model. 
  Additionally, considering that a content might possess multiple genres (e.g., a fictional action movie containing both fiction and action genres) and the predilection might vary across users as well, we further design a user-specific attention (\texttt{UsAttn}) mechanism for a more fine-grained aggregation of various semantics, so as to improve the utilization of the diversified semantic features.
  Unlike the attention mechanism that only considers the influence of genres in content, \texttt{UsAttn} leverages user-content pairs in the bipartite graph and capably analyzes the attraction of semantic information to a specific user during the prediction.
  Meanwhile, given the massive relevancy among users to the same content and the complication to distinguish one specific user from a content-centered sub-graph, as shown in Fig. \ref{DG}, the aforementioned enhancement in \texttt{UsAttn} cannot be applied into the semantic analysis in our attention-based structural learning module. 
  Instead, it requires some modification from some fresh perspective.
  Specifically, inspired by the preliminary effectiveness of a dot-product-based positional encoding (PE) function in Transformer \cite{vaswani2017attention}, we develop a semantic positional encoding function (\texttt{SPE}), deduced from a Fourier kernel-based method, to improve the effectiveness of incorporating multi-dimensional semantics in structural learning.
  Furthermore, we apply our proposed model to a caching strategy in a multi-tier caching system and conduct extensive simulations to evaluate its superiority. 
  Overall, the main contributions of this paper are summarized as follows.

 \begin{itemize}
  \item To deal with the data sparsity, we propose an STGN to leverage the implicit connections between requested contents and their semantic features from both temporal and structural learning perspectives.
  \item Motivated by the fact that a content usually carries various semantic information, we devise a \texttt{UsAttn} mechanism to exploit potential semantic correlations within the user-content bipartite graph and strengthen the temporal learning.
  \item In order to improve the effectiveness of semantics in structural learning, we incorporate a theoretical-grounded multi-dimensional \texttt{SPE}, derived from Fourier kernel, into the attention-based graph learning module.
  \item Extensive experiments based on a real-world dataset verify the improvement in prediction performance achieved by our STGN model and validate the effectiveness of the STGN model-based proactive caching strategy in terms of the cache hit rate.
 \end{itemize}

 \begin{table}[]
    \renewcommand\arraystretch{1}
    \normalsize
    \centering
    \caption{Major notations used in the paper.}
    \label{n_table}
    \scalebox{0.8}{
    \begin{tabular}{ll}
    \hline
    Notation & Definition \\ \hline
    $u_j$, $i_k$ & User $j$ and content $k$\\
    $\mathcal{U}$, $\mathcal{I}$ & The sets of users and contents\\
    $\textbf{\emph{v}}_{u_j}, \textbf{\emph{v}}_{i_k},\textbf{\emph{e}}_{jk}$ & Raw features of $u_j$, $i_k$ and their edges \\
    $\mathcal{V_U}$, $\mathcal{V_I}$, $\mathcal{E}$  & The raw feature sets of users, contents and their edge \\
    $p_{jk}(\hat{T})$ & Real preferences of $u_j$ for $i_k$ at $\hat{T}$ \\
    $\tilde{p}_{jk}(\hat{T})$ & Predicted preferences of $u_j$ for $i_k$ at $\hat{T}$ \\
    $p_{\text{thre}}$ & Threshold value for judging\\
    $\text{Pop}^{i_k}(\hat{T})$ & Popularity of $i_k$ at $\hat{T}$\\
    $\mathbb{P}^{i_k}(\Delta_P)$ & Popularity of $i_k$ during the cache updating period $\Delta_P$ \\
    $\delta_p$ & Popularity predicting period \\
    $K$ & Maximum number of caching capability \\
    $\mathcal{C}(\Delta_P)$ & Real request set during the update period \\ 
    $\widetilde{\mathcal{C}} (\Delta_P)$ & Predicted popularity set for all contents during $\Delta_P$ \\
    $\widetilde{\mathcal{C}}_K (\Delta_P)$ & Caching set during $\Delta_P$ \\
    $h(\Delta_P)$ & Cache hit rate \\
    $\textbf{{Msg}}_{jk}$ & Message that merges all raw features of an interaction\\
    $\textbf{{Msg}}'_{jk}$ & The semantics-enhanced message \\
    $\textbf{{h}}_{j}(\hat{T})$ & Short-term preference of $u_j$ \\
    $\textbf{{Mem}}_{j}$ & Long-term preference of $u_j$\\
    $\textbf{{Mem}}'_{j}$ & \begin{tabular}[c]{@{}l@{}}The updated long-term preference of $u_j$ concatenated\\ with extra features \end{tabular}\\
    $\textbf{{E}}_{u_j}(\hat{T}), \textbf{{E}}_{i_k}(\hat{T})$ & Final embedding representations for $u_j$ and $i_k$ \\
    $\textbf{{E}}_{jk}$ & User-specific embedding\\
    $\textbf{\emph{s}}_{kN_s}$ &  The $N_s$-th semantic information of $i_k$ \\
    $\textbf{\emph{S}}_{jk}$ & User-specific semantic feature for $u_j$ and $i_k$\\
    $\textbf{\emph{S}}_{k}$ & Aggregated semantic feature for $i_k$\\\hline
    \end{tabular}
    }
    \label{notions}
 \end{table}

  The remainder of this paper is organized as follows. The related works are discussed in Section \ref{sec2}. Then, we introduce system models and formulate the problem in Section \ref{sec3}. We elaborate on the details of the main contributions, the proposed prediction model and its modified versions for effective semantic learning in Section \ref{sec4} and \ref{sec5}. In Section \ref{sec6}, we present the experimental results and discussions. Finally, the conclusion is summarized in Section \ref{sec7}. 
  
  For convenience, we also list the mainly used notations of this paper in Table \ref{notions}.

\section{Related Work}\label{sec2}

  Traditionally, albeit the remarkable portability, the widely-mentioned reactive caching strategies, such as least recently used (LRU) and least frequently used (LFU), only focus on the patterns of local requests, thus failing to handle the unexpected requests \cite{970573}.
  Accordingly, it becomes inevitable to design proactive caching strategies, wherein accurate popularity prediction plays a decisive role \cite{9187344}.
  With the development of artificial intelligence (AI), applying DNN to predict popularity has thrived.
  For instance, in \cite{8172025}, a model based on stacked autoencoders (SAE) is proposed to compute the popularity from the content request sequence.
  In addition, Refs. \cite{9234632, 9846902} use a recurrent neural network (RNN) and its variant LSTM to discover the patterns within the temporal content requests, so as to facilitate popularity-assisted content caching. 
  Nevertheless, due to the insufficient historical data, LSTM or other sequence-based prediction models may fail to predict accurately for those inactive users. 
 
  Furthermore, the user-content interactions constitute a bipartite graph and lay the very foundation for adopting GNN to enhance the learning performance \cite{wu2022graph}. 
  Although GNN has won remarkable achievement in RS \cite{zhou2020graph}, most existing works assume that the underlying graph is static, which does not conform to the real-life \cite{skarding2021foundations}. 
  Consequently, popularity prediction with DGNN has been attracting significant attention. 
  Different from the conventional GNN, a DGNN model is able to jointly learn the structural and temporal patterns of dynamic graphs.
  For example, Ref. \cite{trivedi2019dyrep} proposes a DyRep model to calculate the dynamic graph with a recurrent architecture.
  Learning from the ``positional encoding" of the self-attention mechanism in Transformer \cite{vaswani2017attention}, Ref. \cite{Xu2020Inductive} proposes a ``time encoding function'' to encode the timestamp information for the graph attention network (GAT) \cite{velivckovic2017graph}, which is called the temporal graph attention mechanism (TGAT).
  TGN in \cite{rossi2020temporal} introduces an additional temporal learning module on top of the TGAT for a deeper refinement of the temporal characteristics. 
  Ref. \cite{zhu2022aoi} optimizes the temporal learning module of TGN with an age of information (AoI) based attention mechanism to filter and aggregate fresh historical messages, and realizes satisfactory results in content caching. 
  Nevertheless, when most users in the graph are inactive, it is in general difficult to obtain satisfactory caching performance by deploying existing DGNN models in a straightforward way. 

  To overcome the challenge due to the data sparsity, Ref. \cite{Xu2020Inductive} suggests that it is beneficial to stack more TGAT layers for enlarging the receptive field, but it comes at the expense of massive computation cost.
  On the other hand, some works attempt to solve this problem by supplementing the side information of the bipartite graph (e.g., user social influence \cite{9220908}). 
  Refs. \cite{wang2018ripplenet, 9409651, wang2019explainable, 9354956, liu2021contextualized} propose to introduce a knowledge graph (KG) for incorporating the side information of the requested contents into a static graph model, which leverages the implicit associations among the contents and yields superior prediction performance.
  However, these works ignore the dynamics of the interaction graph, while the KG construction also implies the demand for a significant amount of side information (e.g., the director and release date of the content), which may not be available in many cases.
  Therefore, it is more worthwhile to leverage limited content information (e.g., genre information) with a deeper excavation.
  In that regard, the astonishing development of NLP, such as one-hot, BERT \cite{DBLP:conf/naacl/DevlinCLT19}, and Glove \cite{pennington2014glove}, makes it promising to capture the implicit semantic relations between the words.

  Meanwhile, it is meaningful to develop effective means of computing embeddings, so as to better unveil correlations. 
  As for the temporal learning, it is simple and sufficient to adopt a \texttt{UsAttn}-based mechanism to discover the relationship between multiple genres related to contents and users. 
  However, it becomes troublesome for the structural learning module, considering the massive relevance among users to the same content.
  The illuminated work in Ref. \cite{Xu2020Inductive}, which generalizes the definition of position and encodes the timestamps with a customized PE function, motivates us to treat the semantic information as a special kind of position.
  Therefore, we adopt an \texttt{SPE} to strengthen the association analysis between two semantics-attached content embeddings.
  As a specially designed PE function, it is also inspired by works with learnable approaches to encode positions \cite{DBLP:conf/naacl/DevlinCLT19, kitaev2020reformer}.
  Besides, considering the heavy computational cost and non-uniform decay in different dimensions to encode each dimension independently before the concatenation \cite{li2021learnable, 46840}, in \texttt{SPE}, we treat the multi-dimensional position as a whole and encode it directly by learnable Fourier features \cite{li2021learnable}.
  To our best knowledge, the proposed \texttt{SPE} belongs to the first work to view the multi-dimensional semantic information as the position and encode from the perspective of Fourier features. 
  
\section{System Models and Problem Formulation}\label{sec3}
\subsection{System Models}
\subsubsection{Network Model} 
\begin{figure}[tbp]
  \centering 
  \includegraphics[scale = 0.4]{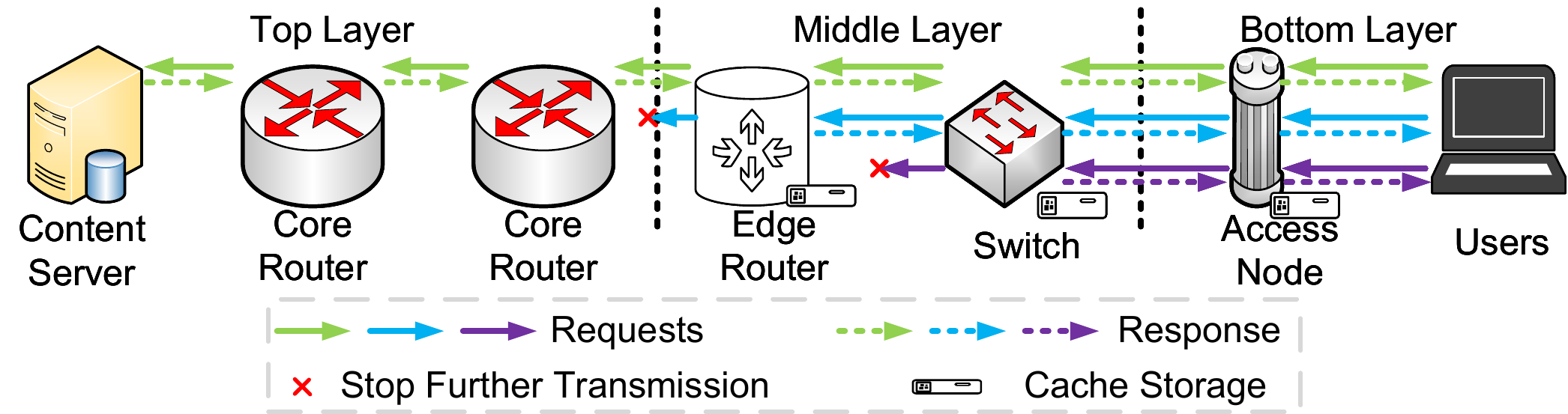}
  \caption{Caching and response in a multi-tier caching system.}  
  \vspace{-1.2em}
  \label{ICN}
 \end{figure}

  \ 
  \newline
  \indent As shown in Fig. \ref{ICN}, we concentrate on a multi-tier caching system, where caches are scattered over the devices close to users, such as the \textit{edge routers}, \textit{switches}, and some \textit{access equipment}. 
  We conceptually simplify the network as a three-layer topology as below.

  \begin{itemize}
      \item \textit{Top Layer} -- It is composed of \textit{core routers}, which are responsible for connecting content providers with other network elements.
      \item \textit{Middle Layer} -- It encompasses \textit{edge routers} and \textit{switches}. In particular, the \textit{switches} usually connect various devices in a network and communicate with the core network through the \textit{edge routers}. And the location of \textit{switches} is lower than the \textit{edge routers}, while they are in the same layer.
      \item \textit{Bottom Layer} -- It consists of \textit{access equipment}, which are deployed to connect users with the \textit{switches}.
  \end{itemize}
  
  In this paper, we primarily take account of the in-network caching capability of the \textit{access equipment}, \textit{switch}, and \textit{edge router}, and denote them as \textit{Tier 1}, \textit{Tier 2}, and \textit{Tier 3}, respectively.
  Moreover, as depicted in Fig. \ref{ICN}, once a copy of the target content is cached at a lower-tier device, the request will be directly responded and no longer be sent to any higher-tier devices.
  
\subsubsection{Request Model}
  \ 
  \\ 
  \indent In this paper, we model the request records in the format of user-content pairs as a graph.
  We denote the set of users as $\mathcal{U} = \{u_0, u_1,..., u_j\}$ and the set of contents as $\mathcal{I} = \{i_0, i_1,..., i_k\}$, where $u_j$ and $i_k$ denote the user $j$ and content $k$, respectively. 
  Furthermore, we allocate raw features for the sets of users and contents with a random initialization method, which are deemed as the input of the predictive model. 
  The raw feature sets of users and contents are denoted as $ \mathcal{V_U} = \{ \textbf{\emph{v}}_{u_0},\textbf{\emph{v}}_{u_1},...,\textbf{\emph{v}}_{u_j}\}$ and $\mathcal{V_I} = \{\textbf{\emph{v}}_{i_0},\textbf{\emph{v}}_{i_1},...,\textbf{\emph{v}}_{i_k}\}$, where $\textbf{\emph{v}}_{u_j}$ and $\textbf{\emph{v}}_{i_k}$ are the vertexes in the dynamic bipartite graph corresponding to $u_j$ and $i_k$, respectively. 
  The interactions, i.e., users requesting contents, can be naturally regarded as the edges, which can be denoted as $\mathcal{E} = \{\textbf{\emph{e}}_{00}, \textbf{\emph{e}}_{01},..., \textbf{\emph{e}}_{jk}\}$. 
  Herein, $\textbf{\emph{e}}_{jk}$ represents the vector of interactions between $u_j$ and $i_k$, and reflects the user-behavior type (e.g., watching videos or listening to music).
  
  Next, we formulate the the evolving interactions as a dynamic graph using a set of quadruples, $\mathcal{G}=\{(\textbf{\emph{v}}_{u_0}, \textbf{\emph{v}}_{i_0},\textbf{\emph{e}}_{00}, T_0), ...,(\textbf{\emph{v}}_{u_j},\textbf{\emph{v}}_{i_k},\textbf{\emph{e}}_{jk}, T_n)\}$, where $T_n$ denotes the timestamp of the $n$-th interaction\footnote{Notably, for simplicity of representation, we omit the superscript $j$ and $k$ of the vertexes $T_n^{jk}$, which represents the $n$-th interaction that happens between $u_j$ and $i_k$.}. 
  In addition, we integrate each quadruple into a piece of historical message as the input of our DGNN model. 
  For instance, the interaction occurred at $T_n$ between $u_j$ and $i_k$ is formulated as $\textbf{{Msg}}_{jk} =[\textbf{\emph{v}}_{u_j} || \textbf{\emph{v}}_{i_k} || \textbf{\emph{e}}_{jk} || T_n]$, where $||$ is the concatenation operator. 
  Moreover, as presented before, content may contain various semantic genres, and we need to encode all $N_s$ genres of content $i_k$ with the NLP methods for fully utilizing the inherent semantic characteristics in the subsequent prediction. 
  Correspondingly, the encoded semantic features are represented as $\mathcal{{S}}_{k} = \{\textbf{\emph{s}}_{k1}, ..., \textbf{\emph{s}}_{kN_s}\}$.

\subsection{Problem Formulation}
  \label{cache}

  In this paper, we evaluate the performance of our STGN model in the caching task with cache hit rate.
  Considering the existence of a maximum number of caching items $K$, only the top-$K$ contents $\tilde{\mathcal{C}}_K(\Delta_P)$ in a popularity ranking list $\widetilde{\mathcal{C}} (\Delta_P)$ are cached during the cache updating period $\Delta_P$.  
  Given the real request set is $\mathcal{C}(\Delta_P)$, we calculate the hit rate during the cache updating period $\Delta_P$ with 
  \begin{equation}
  h(\Delta_P) = \frac{\textbf{I}(\mathcal{C}(\Delta_P), \tilde{\mathcal{C}}_K(\Delta_P))}{\textbf{I}(\mathcal{C}(\Delta_P), \mathcal{C}(\Delta_P))},
  \label{hit_rate}
  \end{equation}
  where $\textbf{I}(\mathcal{X}, \mathcal{Y})$ represents the hit number for the element in $\mathcal{Y}$ to $\mathcal{X}$.

  In line with the previous analysis, to maximize cache hit rate $h(\Delta_P)$, the more popular contents should be cached at the devices closer to the users \cite{ayoub2018energy}. Therefore, it becomes essential to know the popularity of each content and obtain the popularity ranking list $\widetilde{\mathcal{C}} (\Delta_P)$ in advance.
  We assume the list is sorted based on the popularity combining outcomes from several time slots $\delta_p \ll \Delta_P$.
  Therefore, the overall popularity of $i_k$ during the update period $\Delta_P$ can be formulated as 
  \begin{equation}\label{popo}
    \mathbb{P}^{i_k}(\Delta_P) = \sum\nolimits_{n_\delta \in N_\delta}\text{Pop}^{i_k}(n_\delta \times \delta_p),\  \forall i_k \in\mathcal{I},
  \end{equation}
  where $ N_\delta = \{0, 1,\cdots, \left\lfloor \frac{\Delta_P}{\delta_p}\right\rfloor\}$, $\lfloor \rfloor $ is a floor operator, and $\text{Pop}^{i_k}(\hat{T})$ represents the popularity of $i_k$ at the future time $\hat{T} = n_\delta \times \delta_p$.
  Consequently, the popularity ranking list can be obtained to guide the content caching task.
  Notably, in order to distinguish the contents with the same popularity, we decide their prioritization consistent with LRU.

  Obviously, the popularity $\text{Pop}^{i_k}(\hat{T})$ is indispensable, and it can be obtained by gathering the preferences of all users \cite{8425746}, which we calculate as
  \begin{equation}\label{popo}
      \text{Pop}^{i_k}(\hat{T}) = \sum\nolimits_{j}\textbf{1}\left({p}_{jk}(\hat{T}) > p_{\text{thre}}\right),\  \forall i_k \in\mathcal{I},
  \end{equation}
  where ${p}_{jk}(\hat{T})$ indicates the real preference of $u_j$ for $i_k$ at $\hat{T}$, $p_{\text{thre}}$ is the threshold value for judging emergence of such a request, and $\textbf{1}(\zeta)$ is an indicator function that only equals $1$ if the condition $\zeta$ is satisfied.
  
  As real preference $p_{jk}$\footnote{Notably, for simplicity of representation, we omit the $\hat{T}$ of the $p_{jk}(\hat{T})$, $\tilde{p}_{jk}(\hat{T})$, the embeddings $\textbf{{E}}_{u_j}(\hat{T})$ and $\textbf{{E}}_{i_k}(\hat{T})$ in the following equations.} is unknown apriori, we aim to calculate a predicted result $\tilde{p}_{jk}$ with the embeddings of $u_j$ and $i_k$ at $\hat{T}$, namely $\textbf{{E}}_{u_j}(\hat{T})$ and $\textbf{{E}}_{i_k}(\hat{T})$. That is,
 \begin{equation}
      \label{corr}
      \tilde{p}_{jk}(\hat{T})= {F}\left(\textbf{{E}}_{u_j}(\hat{T}), \textbf{{E}}_{i_k}(\hat{T})\right),
 \end{equation}
  where a multi-layer perceptron (MLP) can be adopted to realize the function $F(\cdot)$. 
  In this regard, our target in \eqref{hit_rate} converts to generating feasible representations with the predictive model from the dynamic interaction graph, so as to minimize the binary cross entropy loss (BCELoss) between the real preference, $p_{jk}$, and the predicted one, $\tilde{p}_{jk}$, $\forall u_j \in \mathcal{U}$, $i_k \in \mathcal{I}$,
 \begin{equation}
    \mathcal{L} = -\sum_{u_j, i_k} \left(p_{jk}\log(\tilde{p}_{jk}) + (1 - p_{jk})\log(1 - \tilde{p}_{jk})\right).
 \end{equation}

\section{Semantics-enhanced Temporal Graph Network}\label{sec4}
  In this section, we focus on the design of the STGN, so as to obtain the desired embedding representations, $\textbf{{E}}_{u_j}(\hat{T})$ and $\textbf{{E}}_{i_k}(\hat{T})$, $\forall u_j \in \mathcal{U}$, $i_k \in \mathcal{I}$, from a sparse dataset. 

 \begin{figure*}[htbp]
    \centering 
    \includegraphics[scale =0.62]{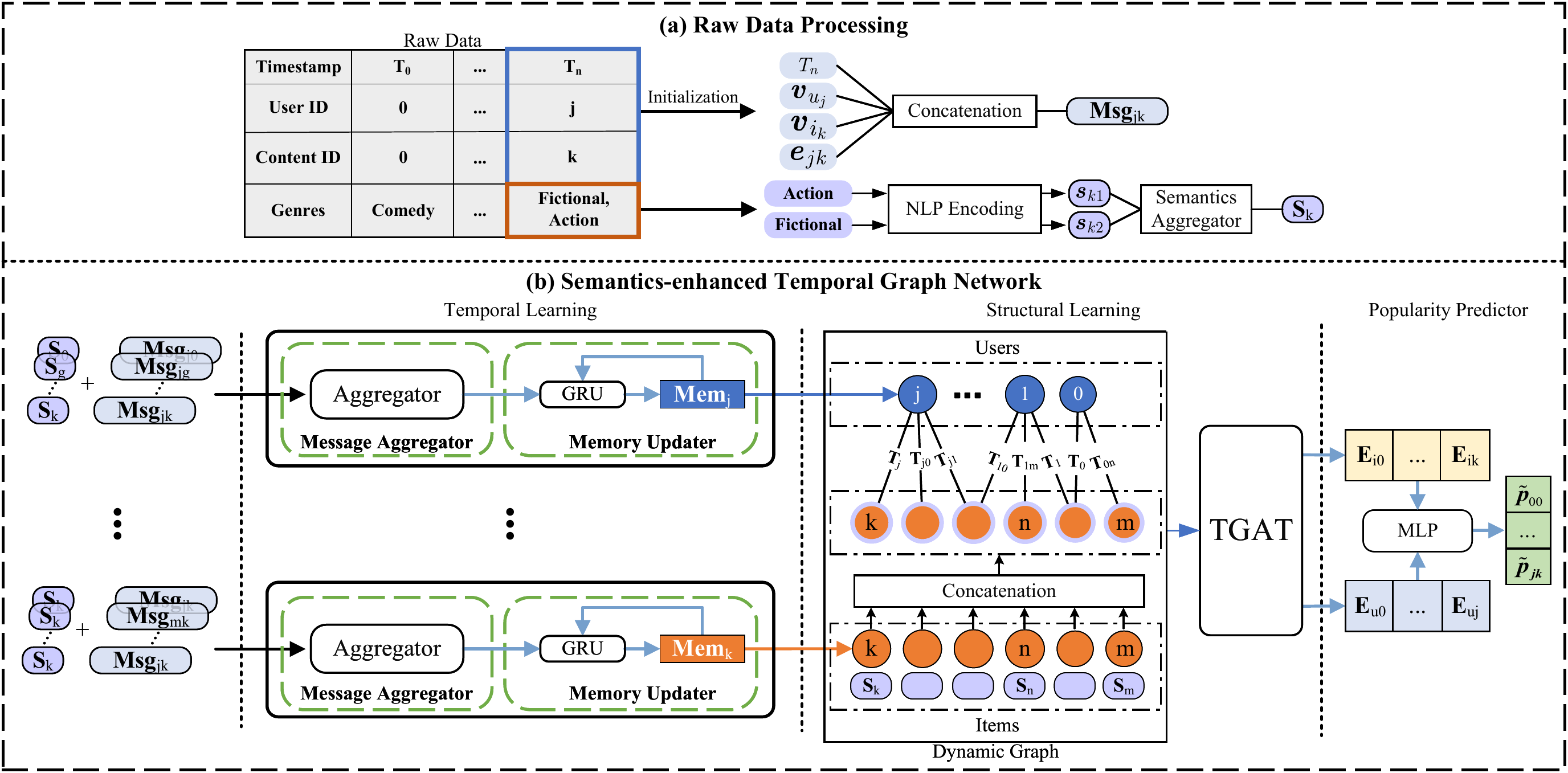}
    \caption{An illustration of \texttt{M2-STGN}, a TGN model enhanced with semantics in both temporal and structural learning.}
    \vspace{-1.2em}
    \label{TGN_s}
 \end{figure*}

\subsection{The Conventional TGN}\label{4B1}
  According to the different roles in prediction, the conventional TGN model is come down to two prime segments, including the temporal learning module and the structural learning module. 

\subsubsection{Temporal Learning Module}\

  The temporal learning module, which consists of a message aggregator and a memory updater, is adopted to compress a user's historical messages into a refined representation. 
  Specifically, the message aggregator leverages several fresh historical messages of $u_j$ before the prediction time $\hat{T}$ to obtain a compressed feature $\textbf{\emph{h}}_j(\hat{T})$. 
  Thus, $\textbf{\emph{h}}_j(\hat{T})$ can also be deemed as a feature that is able to represent the short-term preference of $u_j$, which can be formulated as 
  \begin{equation}
      \textbf{\emph{h}}_j(\hat{T}) = \text{Agg}\left(\textbf{{Msg}}_{j0}, ..., \textbf{{Msg}}_{jk}\right),
      \label{aggre}
  \end{equation}    
  where $\text{Agg}(\cdot)$ is a filtering and aggregation function that can be implemented diversely. 
  In the remainder of this paper, we primarily consider three approaches, including filtering the latest message, using the mean value of all messages \cite{rossi2020temporal}, and an attention-based weighted summation of limited fresh messages with an AoI filter \cite{zhu2022aoi}. 
  We denote them as \texttt{TGN-L}, \texttt{TGN-M}, and \texttt{TGN-A}, respectively.
  
  Subsequently, in order to acquire a much more representative temporal feature, a memory updater is adopted to update the long-term preference $\textbf{{Mem}}_j$ based on the compressed short-term preference $\textbf{\emph{h}}_j(\hat{T})$.
  In order to realize the updater, a learnable function, such as LSTM or the gated recurrent unit (GRU), is necessary.
  Here, considering the advantage in convergence speed \cite{chung2014empirical}, we complete the update procedure with a GRU, which is mathematically formulated as
  \begin{equation}
  \label{gru}
  \begin{aligned}
  &\textbf{{Mem}}_j \leftarrow\textbf{\emph{Z}}\cdot\textbf{\emph{H}} +\left(1-\textbf{\emph{Z}}\right)\cdot \textbf{{Mem}}_j,\\
  &\textbf{\emph{Z}}=\sigma\left(\textbf{\emph{h}}_j(\hat{T})\textbf{\emph{W}}_{hZ}+\textbf{{Mem}}_j\textbf{\emph{W}}_{MZ}+\textbf{\emph{b}}_Z\right),\\
  &\textbf{\emph{H}}=\text{tanh}\left(\textbf{\emph{h}}_j(\hat{T})\textbf{\emph{W}}_{hH}+\left(\textbf{\emph{F}}\cdot\textbf{{Mem}}_j\right)\textbf{\emph{W}}_{MH}+\textbf{\emph{b}}_H\right),\\
  &\textbf{\emph{F}}=\sigma\left(\textbf{\emph{h}}_j(\hat{T})\textbf{\emph{W}}_{hF}+\textbf{{Mem}}_j\textbf{\emph{W}}_{MF}+\textbf{\emph{b}}_F\right),
  \end{aligned}
  \end{equation}
  where $\textbf{\emph{W}}_{hZ}$, $\textbf{\emph{W}}_{hF}$, $\textbf{\emph{W}}_{hH}$, $\textbf{\emph{W}}_{MZ}$, $\textbf{\emph{W}}_{MF}$ and $\textbf{\emph{W}}_{MH}$ denote the trainable weights, while $\textbf{\emph{b}}_Z$, $\textbf{\emph{b}}_F$ and $\textbf{\emph{b}}_H$ are the learnable bias values of the GRU. $\sigma(\cdot)$ and $\rm{tanh}(\cdot)$ are the activation functions.

\subsubsection{Structural Learning Module}\ 
  
  The structural learning module aims to generate embeddings for future prediction.
  Especially, it is also responsible for keeping the representations of the inactive users up-to-date by exchanging features among neighbors in the graph. 
  Obviously, the timestamp of each interaction also plays a vital role in the mapping procedure. 
  Therefore, we adopt a TGAT model \cite{Xu2020Inductive} to accomplish this unconventional structural learning.
  Notably, the TGAT mechanism is a module that deploys a learnable time encoding function on the basis of a classical GAT module \cite{velivckovic2017graph}. 
  In particular, the specially designed time encoding function is formulated as
 \begin{equation}\label{encodet}
  \Phi _{d_T}(\Delta_t) = \sqrt{\frac{1}{d_T}}[\cos(\omega _1\Delta_t), ..., \cos(\omega _{d_T}\Delta_t)]^T,
 \end{equation}
  where $\omega_1, \omega_2, ...$ and $\omega_{d_T}$ are the trainable parameters, $\Delta_t$ denotes the time slot between the interaction-occurring time $T_n$ and the time to predict $\hat{T}$, (i.e., $\Delta_t = \hat{T}-T_n$).
  $d_T$ is the dimension number of the desired time encoding.
  
  Then, the encoded time features are concatenated to the output of the temporal learning module $\textbf{{Mem}}_j$ as the input for the structural learning. 
 \begin{equation}
      \textbf{{Mem}}'_{j} = [\textbf{{Mem}}_j || \Phi _{d_T}(0)],\label{mem_1}
 \end{equation}
  where \eqref{mem_1} supplements the updated long-term preference $\textbf{{Mem}}'_{j}$ of $u_j$ with the time feature.
  It is noteworthy that $u_j$ is the center node of a user-centered sub-graph that we want to learn, so we define $\Delta_t=0$ for its prediction. 
  As for $u_j$'s neighbor $k \in \mathcal{N}_j $, its modified preference term $\textbf{{Mem}}'_{k}$ is formulated as 
 \begin{equation}
      \textbf{{Mem}}'_{k} = [\textbf{{Mem}}_k || \Phi _{d_T}(\Delta_{t_k})], \forall k \in \mathcal{N}_j, \label{mem_2}
 \end{equation}
  Notably, $\textbf{{Mem}}'_{j}$ and $\textbf{{Mem}}'_{k}$ are the inputs for the structural learning module.
  As depicted in Fig. \ref{TGN_s}, the GAT architecture \cite{velivckovic2017graph} is the paramount part of a TGAT layer to achieve the structural learning of $u_{j}$'s dynamic sub-graph, and it can be encapsulated as
 \begin{equation}
  \label{TGAT}
  \textbf{{E}}_{u_j}(\hat{T}) = \text{GAT}(\textbf{{Mem}}'_{j}, \textbf{{Mem}}'_{\mathcal{N}_j}),
 \end{equation}
  Similarly, we can generate the embedding $\textbf{{E}}_{i_k}(\hat{T})$ from the content-centered sub-graph of $i_k$ with
 \begin{equation}
      \begin{aligned}
      &\textbf{{Mem}}'_{k} = [\textbf{{Mem}}_k || \Phi _{d_T}(0)],\\ 
      &\textbf{{Mem}}'_{j} = [\textbf{{Mem}}_j || \Phi _{d_T}(\Delta_{t_j})], \forall j \in \mathcal{N}_k,\\
      &\textbf{{E}}_{i_k}(\hat{T}) = \text{GAT}(\textbf{{Mem}}'_{k}, \textbf{{Mem}}'_{\mathcal{N}_k}),
      \end{aligned}
 \end{equation}
  Note that $\textbf{{E}}_{u_j}(\hat{T})$ and $\textbf{{E}}_{i_k}(\hat{T})$ are utilized as the input of the prediction module in \eqref{corr}. 
  Furthermore, the stacking of multiple TGAT layers can leverage more hidden information within the graph by aggregating multi-hop neighbors. 
  But the enlargement of receptive field also implies greater computational complexity \cite{Xu2020Inductive}. 
  Thus, we only investigate the performance with a one-layer TGAT to shorten the training session in our simulations.

\subsection{Semantic Enhancement for TGN}\label{4B2}
  Essentially, the temporal learning module in TGN can be deemed as a procedure for refining the commonality from the temporal perspective. 
  However, the randomly initialized raw features make it complicated to accurately extract and analyze the patterns, especially for a sparse dataset. 
  Consequently, we resort to supplementing the raw input with semantic information, so as to improve the abilities of reasoning and interpretability of our model by extracting the implicit semantic correlations among the contents. 
  
  We use some pre-trained NLP models, such as one-hot, BERT \cite{DBLP:conf/naacl/DevlinCLT19}, and Glove \cite{pennington2014glove}, to encode the content genre information as semantic messages, $\mathcal{{S}}_{k} = \{\textbf{\emph{s}}_{k1}, ..., \textbf{\emph{s}}_{kN_s}\}$. 
  For the sake of simplicity, we adopt the summation as a semantic aggregator to generate the aggregated feature $\textbf{\emph{S}}_{k}$ from $\mathcal{{S}}_{k}$, which is then incorporated into the raw message as shown in Fig. \ref{TGN_s}(b). 
  In specific,
 \begin{align}
       &\textbf{\emph{S}}_{k} = \sum\nolimits_{ n \in N_s} \sigma(\textbf{\emph{W}}_{s}\textbf{\emph{s}}_{kn}+\textbf{\emph{b}}_s), \label{sum_a}\\
       &\textbf{{Msg}}'_{jk}=\sigma(\textbf{\emph{W}}^t_{1}\textbf{{Msg}}_{jk}+ \textbf{\emph{W}}^t_{2}\textbf{\emph{S}}_k), \label{sum_b}
 \end{align}
  where $\textbf{\emph{W}}_{s}$, $\textbf{\emph{b}}_s$, $\textbf{\emph{W}}^t_{1}$ and $ \textbf{\emph{W}}^t_{2}$ are the trainable parameters to enhance the semantic features, while $\textbf{{Msg}}'_{jk}$ is the desired semantics-enhanced historical message in \eqref{aggre}. As $\textbf{{Msg}}'_{jk}$ can be directly applied to enhance the temporal learning by replacing $\textbf{{Msg}}_{jk}$ in \eqref{aggre}, we regard such an approach as the semantics-enhanced TGN in a temporal manner, and denote it as \texttt{M1-STGN}.
  
  As we discussed above, although the fresh features for inactive users can be located with the help of the graph structure, the performance still suffers from data sparsity. 
  To address this issue, we further attach the semantic features to the input of the structural learning module, establishing implicit semantic pathways for the dynamic graph from the semantic sphere. 
  In our experiments, we also discover that concatenation outperforms summation for merging semantics in the structural learning module. 
  Then, \eqref{mem_2} is further modified as
 \begin{equation}
  \label{TGAT_0}
  \textbf{{Mem}}'_k = [\textbf{{Mem}}_k || \textbf{\emph{S}}_k || \Phi _{d_T}(\Delta_{t_k})], \forall k \in \mathcal{N}_j,
 \end{equation}
  where $\textbf{\emph{S}}_k$ is calculated by \eqref{sum_a} as well. Similarly, we use \texttt{M2-STGN} to represent the TGN model that is further facilitated by the structural learning with semantics.

\section{Effective Semantics-enhanced Temporal Graph Network}\label{sec5} 
  Although semantic aggregation can be easily achieved with the aforementioned frameworks, their utilization of semantics is still coarse.
  Specifically, the summation semantic aggregator doesn't distinguish the impact of different semantics from the same content on different users, while the concatenation in \eqref{TGAT_0} may be oversimplified to compute proper attention coefficients. 
  Thus, we propose two novel methods to utilize the semantics with high proficiency.

\subsection{User-specific Attention Mechanism for Semantic Aggregation}

 \begin{algorithm}[tbp]
  \caption{The preference prediction with STGN}
  \label{alg:algorithm1}
  \begin{algorithmic}[1]
  \REQUIRE Request dataset and pre-trained NLP model;
  \ENSURE The representations of $u_j$ and $i_k$, (i.e., $\textbf{\emph{E}}^u_{j}$ and $\textbf{\emph{E}}^i_{k}$) and the preference between $u_j$ and $i_k$.\
  \STATE {Initialize the raw data and the parameters for the whole network and encode contents' semantics information with pre-train NLP model.} \
  \STATE {Divide the raw data into several mini batches;}\
  \FOR {each $\rm{batch}(\textbf{\emph{v}}_{u_j}, \textbf{\emph{v}}_{i_k}, \textbf{\emph{e}}_{ui}, t, \mathcal{{S}}_{k}) \in training\  dataset$} 
  \STATE {$\dot{n}\gets \rm{Sample\ negatives}$;} 
  \IF{aggregate $\mathcal{{S}}_{k}$ by summation}
    \STATE {Aggregate $\mathcal{{S}}_{k}$ to compute $\textbf{\emph{S}}_k$ as \eqref{sum_a};}
  \ELSE
    \STATE{Obtain user-specific embedding $\textbf{{E}}_{jk}$ with \eqref{usa_b};}
    \STATE{Aggregate $\mathcal{{S}}_{k}$ and $\textbf{{E}}_{jk}$ with \eqref{usa_a} and \eqref{at_s};}
  \ENDIF
  \STATE {Concatenate the message $\textbf{{Msg}}_{jk}$ with the aggregated semantics in \eqref{sum_b};}
  \STATE {Filter and aggregate historical messages in \eqref{aggre} to obtain short-term preference $\textbf{\emph{h}}_j(\hat{T})$;}
  \STATE {Update long-term preference $\textbf{{Mem}}_{j}$ with $\textbf{\emph{h}}_j(\hat{T})$ in \eqref{gru};}
  \STATE {Encode the time slot $\Delta_{t}$ with \eqref{encodet} for all nodes;}
  \IF{semantic positional encoding}
    \STATE{Encode the summarized semantics with \eqref{pe_1} and \eqref{pe_2};}
  \ENDIF
  \STATE {Incorporate the encrypted time and semantics features into the updated lone-term preference $\textbf{{Mem}}'_j$;}
  \STATE {Obtain $\textbf{{E}}^u_{j}(T_p)$ and $\textbf{{E}}^i_{k}(T_p)$ through the TGAT module;}
  \STATE{Predict the preference between users and content with \eqref{corr};}
  \STATE {Optimize with $\rm{BCELoss}(\cdot)$;}
  \ENDFOR
  \end{algorithmic}
 \end{algorithm}

\begin{figure}[tbp]
    \centering 
    \includegraphics[scale =0.48]{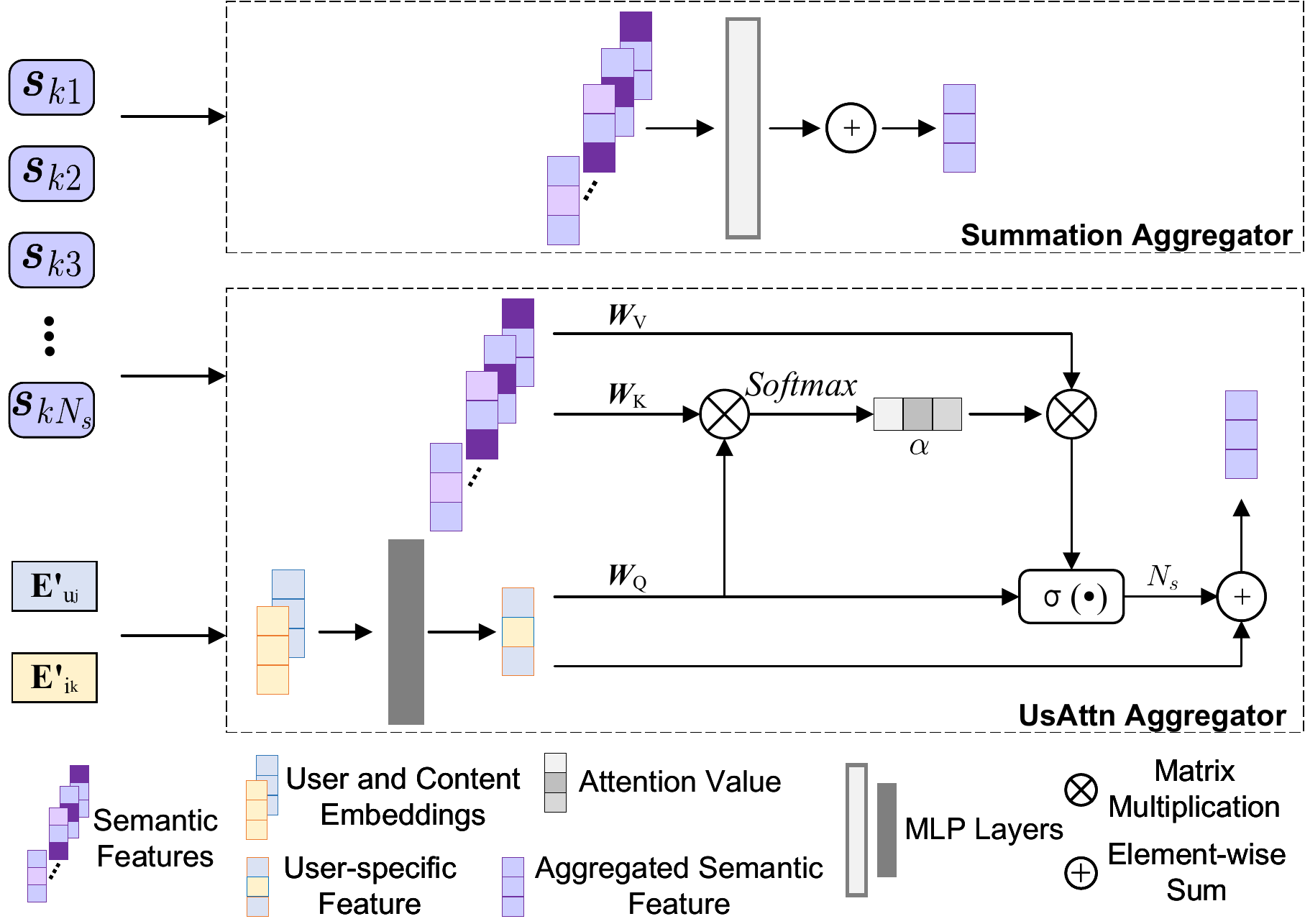}
    \caption{The illustration of the semantics aggregators. (Upper: Summation Aggregator; Lower: UsAttn Aggregator)}  
    \vspace{-1.2em}
    \label{USATGN_s}
 \end{figure}

  In order to aggregate multiple semantics fine-grainedly, we can modify the semantic aggregator with an attention mechanism that calculates attention coefficients by analyzing the influence of different genres on the same content.
  However, it ignores the impact from users, which is also paramount in prediction.
  Thus, we adopt a \texttt{UsAttn} mechanism to aggregate the multiple semantics, as shown in Fig. \ref{USATGN_s}. 
  For different users, this mechanism enables the computation of different attention scores for the diverse semantics of the same content and then generates user-specific semantic features.
  Mathematically, for each content and user, \eqref{sum_a} is reformulated as a linear weighted summation of $N_s$ semantics of content $i_k$, where the weights are calculated by the attention mechanism,
 \begin{equation}
    \begin{aligned}
      &\textbf{\emph{S}}_{jk} = \sigma\left(\sum\nolimits_{n\in N_s} \alpha_{jn}\textbf{\emph{s}}_{kn}\textbf{\emph{W}}_{Vn}\right),
      \\
      &\alpha_{jn} = \frac{\exp(\textbf{{E}}_{jk}\textbf{\emph{W}}_Q\cdot\textbf{\emph{s}}_{kn}\textbf{\emph{W}}_{Kn})} {\sum_{m = 1}^{N_s} \exp(\textbf{{E}}_{jk}\textbf{\emph{W}}_Q\cdot\textbf{\emph{s}}_{km}\textbf{\emph{W}}_{Km})}, 
      \label{usa_a}
   \end{aligned} 
 \end{equation}
  where $\textbf{\emph{W}}_{Kn}$, $\textbf{\emph{W}}_{Q}$ and $\textbf{\emph{W}}_{Vn}$ are the trainable parameters, and $\alpha_{jn}$ is the attention coefficient of the $n$-th semantic message of the content.
  Especially, $\textbf{{E}}_{jk}$ is a user-specific embedding, after which we believe the weight calculation will be forced to account for the embeddings of both $u_j$ and $i_k$. 
  Accordingly, we define it as 
 \begin{equation}
    \textbf{{E}}_{jk} = \text{LeakyReLu}(\textbf{\emph{W}}_{u}\textbf{{E}}'_{u_j} +\textbf{\emph{W}}_i\textbf{{E}}'_{i_k}+\textbf{\emph{b}}_{ui}),
      \label{usa_b}
 \end{equation}
  where $\textbf{\emph{W}}_{u}$, $\textbf{\emph{W}}_{i}$ and $\textbf{\emph{b}}_{ui}$ are the trainable parameters, while $\textbf{\emph{E}}'_{u_j}$ and $\textbf{{E}}'_{i_k}$\footnote{For simplicity, we omit the time information $\hat{T}'$ of the last prediction in $\textbf{{E}}'_{u_j}(\hat{T}')$ and $\textbf{{E}}'_{i_k}(\hat{T}')$.} are the results generated in the last prediction or the initialization values for the first round prediction of $u_j$ and $i_k$, respectively.
  
  Moreover, the stacking of multiple DNN layers possibly results in the over-smoothing issue.
  In this regard, we further leverage the skip-connection in Transformer \cite{vaswani2017attention} to avoid this issue and improve the overall performance. Specifically, for each piece of historical message that happened between $u_j$ and $i_k$, the aggregated semantics is denoted as
 \begin{equation}
    \textbf{\emph{S}}_k = N_s\cdot\textbf{\emph{S}}_{jk} + \textbf{\emph{E}}_{jk},
    \label{at_s}
 \end{equation}
  which is the desired representation that we use in \eqref{sum_b}, so as to further optimize \texttt{M1-STGN} or the temporal learning module of \texttt{M2-STGN}.
  
\subsection{Semantic Positional Encoding for Structural Learning}\label{SPE}
  For different user-content pairs, the proposed \texttt{UsAttn} mechanism allocates different attention weights for content's diverse semantic messages in temporal learning.
  However, it is hard to generalize this mechanism into the structural learning module.
  Specifically, the structural learning for $i_k$ is generally conducted by calculating a sub-graph centering around $i_k$, where all its neighbors are users.
  In other words, it is elusive for us to determine a specific user before enhancing the structural learning with \texttt{UsAttn}.
  This dilemma motivates us to find another method to improve semantics utilization in structural learning.
  Inspired by the positional encoding in Transformer and the time encoding in TGAT, as a Transformer-alike module, it might be also feasible to treat the vertexes' semantic features, generated by \eqref{sum_a}, as the locators in semantic sphere during the structural learning.
  Therefore, we ameliorate the model with a specially designed positional encoding function to reinforce the attention effectiveness.
  To extract useful characteristics from the multi-dimensional semantic position $\textbf{\emph{S}}_{k}$, calculated by \eqref{sum_a}, we adopt a learnable Fourier features positional encoding function, which is derived in Appendix and can be mathematically formulated as follows,
 \begin{equation}
    \textbf{\emph{R}}_{k} = \frac{1}{\sqrt{D_h}}[\cos \textbf{\emph{W}}_{p}\phi_1(\textbf{\emph{S}}_{k}) || \sin \textbf{\emph{W}}_{p}\phi_1(\textbf{\emph{S}}_{k})]^T, 
    \label{pe_1}
 \end{equation}
  where $\phi_1(\cdot)$ is an MLP layer to enhance the semantic features, and $D_h$ is the dimension of the hidden layer. 
  Notably, the initialized $\textbf{\emph{W}}_{p}$ is drawn from a normal distribution \cite{rahimi2007random}.
  Furthermore, we also discover that an additional feature enhancement with another MLP is beneficial to the final performance,
 \begin{equation}
    \textbf{\emph{S}}_{k} \leftarrow \textbf{\emph{W}}^2_{p}\text{GeLU}(\textbf{\emph{W}}^1_{p}\textbf{\emph{R}}_{k}), 
    \label{pe_2}
 \end{equation}
  where $\textbf{\emph{W}}^1_{p}$ and $\textbf{\emph{W}}^2_{p}$ is the trainable weights, and $\text{GeLU}(\cdot)$ is an activation function that is widely adopted in NLP tasks \cite{li2021learnable}. 

  After the calculation with \eqref{pe_1} and \eqref{pe_2}, we concatenate the encoded semantic positional embeddings into the input of TGAT layer, as done in \eqref{TGAT_0}. 
  Finally, we summarize all variants of our STGN model in Algorithm \ref{alg:algorithm1} and highlight their key differences in temporal and structural learning modules in Table \ref{STGN_variants}.

 \begin{table}[tbp]
    \centering
    \caption{The variants of our proposed STGN model, where Sum and \texttt{UsAttn} are the semantic aggregators while \texttt{SPE} is the extra positional encoding for the graph attention module. }
    \label{STGN_variants}
    \resizebox{0.45\textwidth}{14mm}{
    \begin{tabular}{c|cc}
    \hline
     & Temporal Learning & Structural Learning \\ \hline
    \texttt{M1-STGN} & \texttt{Sum} & - \\
    \texttt{M2-STGN} & \texttt{Sum} & \texttt{Sum} \\
    \texttt{M1-STGN+U} & \texttt{UsAttn} & - \\
    \texttt{M2-STGN+U} & \texttt{UsAttn} & \texttt{Sum} \\
    \texttt{M2-STGN+SPE} & \texttt{Sum} & \texttt{Sum+SPE} \\
    \texttt{M2-STGN+U+SPE} & \texttt{UsAttn} & \texttt{Sum+SPE} \\ \hline
    \end{tabular}}
 \end{table}

\section{Experimental Results and Discussions}\label{sec6}
  In this section, we present the performance of our proposed models in prediction and caching tasks. We also compare our methods with three state-of-the-art models in processing dynamic graphs, including TGAT \cite{Xu2020Inductive}, DyRep \cite{trivedi2019dyrep}, and TGN \cite{rossi2020temporal}. 
  Besides, in order to analyze the effectiveness of semantic features, we further adopt some widely-accepted NLP methods to encode the genres, i.e., one-hot, BERT \cite{DBLP:conf/naacl/DevlinCLT19}, and Glove \cite{pennington2014glove}. 
  Moreover, experiments with respect to the cache hit rate are also conducted to validate the superiority of our model-based caching methods.

 \begin{table*}[tbp]
    \centering
    \caption{The training time and the performance of predicting content requests in both transductive and inductive tasks. \texttt{TGAT} and \texttt{DyRep} are two state-of-the-art DGNN models. \texttt{TGN-L}, \texttt{TGN-M}, and \texttt{TGN-A} are the conventional TGN model's variants with different message aggregators. The best results are highlighted in \textbf{bold} and the second-best results are highlighted in \underline{underlined}.}
    \label{tab:my-table_1}
    \resizebox{0.98\textwidth}{42mm}{
      \begin{tabular}{cc|cc|cc|c}
        \hline
        \multicolumn{2}{c|}{Metric} & \textit{AUC for Transductive} & \textit{AP for Transductive} & \textit{AUC for Inductive} & \textit{AP for Inductive} & \textit{Training Time} \\ \hline
        \multicolumn{1}{c|}{\multirow{5}{*}{Baseline}} & \texttt{TGAT} & 75.891 & 73.426 & 66.549 & 65.955 & 43.711s \\
        \multicolumn{1}{c|}{} & \texttt{DyRep} & 84.027 & 84.096 & 76.162 & 77.562 & 20.116s \\
        \multicolumn{1}{c|}{} & \texttt{TGN-L} & 85.299 & 83.824 & 77.285 & 76.995 & 14.602s \\
        \multicolumn{1}{c|}{} & \texttt{TGN-M} & 86.731 & 86.022 & 78.953 & 79.721 & 90.729s \\
        \multicolumn{1}{c|}{} & \texttt{TGN-A} & 90.507 & 90.691 & 83.504 & 84.999 & 158.163s \\ \hline
        \multicolumn{1}{c|}{\multirow{3}{*}{\texttt{M1-STGN}}} & \texttt{M1-STGN-L} & 86.386 & 85.892 & 79.980 & 80.439 & 16.291s \\
        \multicolumn{1}{c|}{} & \texttt{M1-STGN-M} & 88.312 & 88.095 & 82.043 & 82.765 & 91.855s \\
        \multicolumn{1}{c|}{} & \texttt{M1-STGN-A} & 91.210 & 91.337 & 85.247 & 86.175 & 159.239s \\ \hline
        \multicolumn{1}{c|}{\multirow{3}{*}{\texttt{M2-STGN}}} & \texttt{M2-STGN-L} & 87.383 & 86.806 & 81.131 & 81.182 & 17.917s \\
        \multicolumn{1}{c|}{} & \texttt{M2-STGN-M} & 88.649 & 88.558 & 82.805 & 83.461 & 91.309s \\
        \multicolumn{1}{c|}{} & \texttt{M2-STGN-A} & {\ul 91.773} & {\ul 91.953} & {\ul 85.877} & {\ul 87.019} & 158.754s \\ \hline
        \multicolumn{1}{c|}{\multirow{3}{*}{\texttt{M1-STGN+U}}} & \texttt{M1-STGN-L+U} & 89.014 & 88.467 & 83.096 & 83.483 & 16.510s \\
        \multicolumn{1}{c|}{} & \texttt{M1-STGN-M+U} & 89.721 & 89.356 & 83.585 & 84.148 & 92.638s \\
        \multicolumn{1}{c|}{} & \texttt{M1-STGN-A+U} & 91.358 & 91.572 & 85.327 & 86.567 & 158.077s \\ \hline
        \multicolumn{1}{c|}{\multirow{3}{*}{\texttt{M2-STGN+U}}} & \texttt{M2-STGN-L+U} & 89.749 & 89.279 & 84.183 & 84.387 & 18.173s \\
        \multicolumn{1}{c|}{} & \texttt{M2-STGN-M+U} & 90.107 & 89.884 & 84.434 & 84.868 & 91.655s \\
        \multicolumn{1}{c|}{} & \texttt{M2-STGN-A+U} & \textbf{91.846} & \textbf{92.056} & \textbf{86.264} & \textbf{87.279} & 163.363s \\ \hline
        \multicolumn{1}{c|}{\multirow{3}{*}{\texttt{M2-STGN+U+SPE}}} & \texttt{M2-STGN-L+U+SPE} & 90.778 & 90.406 & 84.582 & 85.211 & 19.157s \\
        \multicolumn{1}{c|}{} & \texttt{M2-STGN-M+U+SPE} & 90.951 & 90.641 & 84.542 & 85.181 & 92.235s \\
        \multicolumn{1}{c|}{} & \texttt{M2-STGN-A+U+SPE} & 91.240 & 91.418 & 85.187 & 86.430 & 160.544s \\ \hline
        \end{tabular}}
 \end{table*}

 \begin{table}[]
  \centering
  \caption{The test average precision results of \texttt{M2-STGN+SPE} with different NLP methods for encoding the genre information.}
  \label{table_NLP}
  \resizebox{0.45\textwidth}{20mm}{
  \begin{tabular}{cc|cc}
    \hline
    \multicolumn{2}{c|}{Semantic model} & Transductive & Inductive \\ \hline
    \multicolumn{1}{c|}{\multirow{3}{*}{One-hot}} & \texttt{M2-STGN-L+SPE} & 87.827 & 81.581 \\
    \multicolumn{1}{c|}{} & \texttt{M2-STGN-M+SPE} & 89.730 & 84.002 \\
    \multicolumn{1}{c|}{} & \texttt{M2-STGN-A+SPE} & 90.908 & 85.322 \\ \hline
    \multicolumn{1}{c|}{\multirow{3}{*}{BERT}} & \texttt{M2-STGN-L+SPE} & 87.998 & 82.300 \\
    \multicolumn{1}{c|}{} & \texttt{M2-STGN-M+SPE} & 88.972 & 83.598 \\
    \multicolumn{1}{c|}{} & \texttt{M2-STGN-A+SPE} & 91.287 & 86.427 \\ \hline
    \multicolumn{1}{c|}{\multirow{3}{*}{Glove}} & \texttt{M2-STGN-L+SPE} & 88.397 & 82.447 \\
    \multicolumn{1}{c|}{} & \texttt{M2-STGN-M+SPE} & 89.392 & 83.803 \\
    \multicolumn{1}{c|}{} & \texttt{M2-STGN-A+SPE} & 91.805 & 86.800 \\ \hline
    \end{tabular}}
 \end{table}

\subsection{Experimental Settings}\label{4B}
  \textbf{Dataset:} 
  In this paper, the experiments are carried out with a public dataset, Netflix\footnote{https://www.kaggle.com/datasets/vodclickstream/netflix-audience-behaviour-uk-movies}, which records a set of user behaviors on Netflix in UK. 
  Notably, there are many insignificant historical messages, such as some users only request once or watch the content for an extremely short period. 
  The burst behavior is hard to be predicted accurately, and it may mislead other predictions as well.
  Therefore, we select those users who have more than 4 requests and view each requested content for more than 3 minutes as the valid input for the prediction. 
  The dataset we actually adopt includes 86,889 interactions, which involve 11,254 different users and 4,057 pieces of content. 
  The number of interactions is less than the dataset\footnote{The dataset used in Ref. \cite{zhu2022aoi} involves 5,763 users and 56 contents, which consists of 175,856 interactions.} used in Ref. \cite{zhu2022aoi}, while the numbers of users and contents are larger, making the Netflix even much sparser.
  Afterwards, we perform a 60\%-20\%-20\% chronological split of the dataset for training, validation, and testing, respectively. 

 \begin{figure}[tbp]
    \centering
      \subfloat[1 TGAT Layer]{\includegraphics[width = 0.43\textwidth]{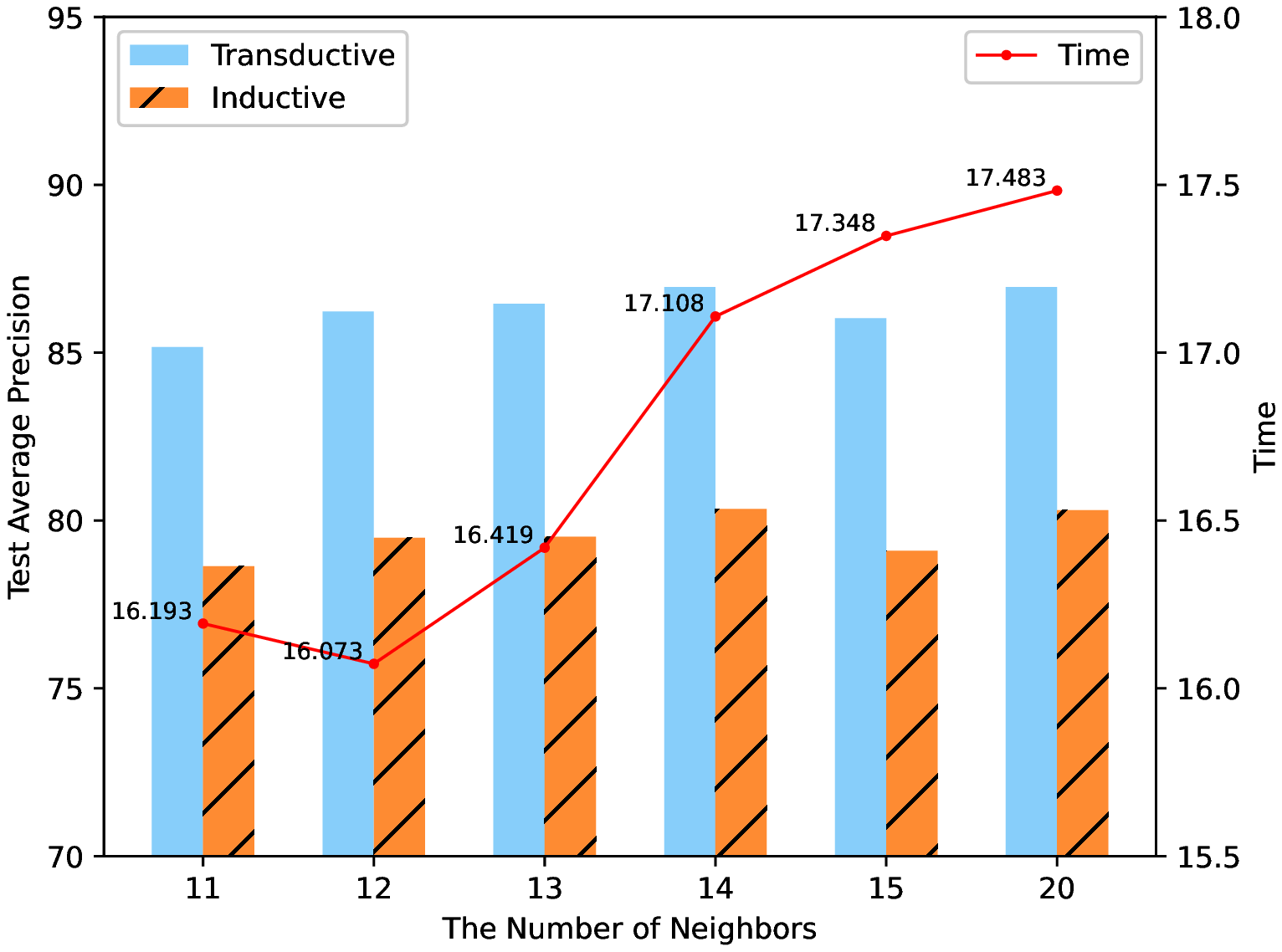}}
      \hfill 
      \subfloat[2 TGAT Layers]{\includegraphics[width = 0.43\textwidth]{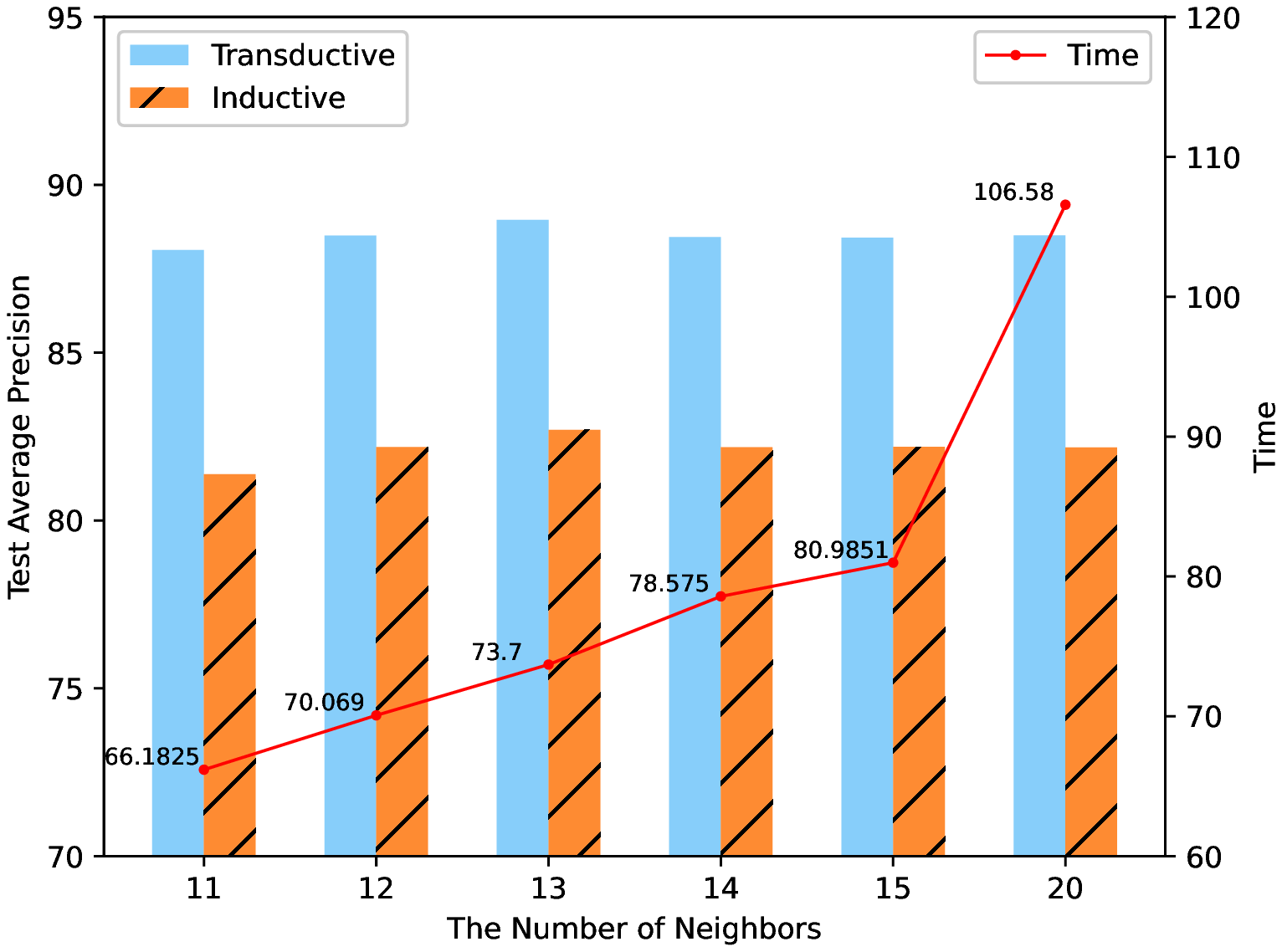}}
    \caption{The test average precision and training time of \texttt{TGN-L} with different numbers of aggregating neighbors and TGAT layers.}
    \vspace{-1.2em}
    \label{fig:layer}
 \end{figure}

 \textbf{Evaluation Tasks and Training Configuration:}
  To verify the effectiveness of our proposed models, we compare our models with several state-of-the-art models, including \texttt{TGAT} \cite{Xu2020Inductive}, \texttt{DyRep} \cite{trivedi2019dyrep}, \texttt{TGN} \cite{rossi2020temporal} and its variants (i.e., \texttt{TGN-L}, \texttt{TGN-M} and \texttt{TGN-A}).
  Notably, the receptive field for the graph in GNN is proportional to the number of GNN layers $l$ and the neighbors $N_n$ in each sub-graph.
  As Ref. \cite{Xu2020Inductive} suggests, we set $l = 2$ and $N_n = 10$ in \texttt{TGAT}, while $l=1$ and $N_n = 10$ in \texttt{TGN}.
  Besides, we also compare our models to the TGN model with larger receptive field, i.e., $l = 1,2$ and $N_n = 11, 12, 13, 14, 15, 20$.

  Furthermore, as for pre-trained NLP models,
  Glove \cite{pennington2014glove} is conducted by global word-to-word occurrence statistics from a large corpus, while BERT \cite{DBLP:conf/naacl/DevlinCLT19} is a neural network model based on 12-layer Transformer.
  The dimension numbers of their output embeddings are fixed (50 for Glove and 738 for BERT).
  To reduce the computational cost, we also adopt an MLP to compress the representations of BERT.
  Besides, Ref. \cite{jawahar2019does} also discovers that the outputs from the $6$-th to $10$-th layers outperform in semantic tasks.
  Therefore, we average the embeddings from the $6$-th to $10$-th layers to investigate the performance.

  Moreover, we conduct experiments under two tasks, i.e., transductive task and inductive task. 
  Different from the transductive task, the validation set and test set in an inductive task may contain some vertices that have not been observed by models during the training phase.
  For both tasks, we adopt the \textit{average precision (AP)} and the \textit{area under the ROC curve (AUC)} as evaluation metrics.

 \begin{figure}[tbp]
    \centering
    \includegraphics[width = 0.43\textwidth]{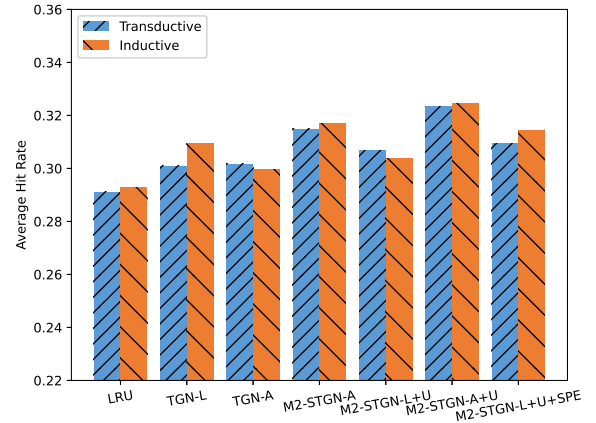}
    \caption{The average hit rate performance of different algorithms in 24 hours.}
    \vspace{-1.2em}
    \label{fig:hitrate}
 \end{figure}

  \textbf{Caching Policy Setting:}
  As depicted in Fig. \ref{ICN}, we configure a multi-layer network architecture. 
  The storage capacity in the device closer to users is typically smaller \cite{ayoub2018energy}. 
  Hence, we assume that the \textit{Tier 1} can store 5 contents, while 7 and 8 contents can be cached at \textit{Tier 2} and \textit{Tier 3}, respectively.

  As for the content caching task, our target is to predict contents' popularity during 24 hours in the test phase.
  We assume that the candidate content set $\mathcal{I}$ is known apriori.
  To make the simulations more practical, we supplement the candidate content set with a noise set, which consists of the contents that have been requested within a 50-hour duration before the prediction starting time.
  Besides, the user set $\mathcal{U}$ of each hour is also assumed to be known in our simulation.
  As for other hyperparameters, we compute the per-hour popularity with $\Delta_P = 1$h and $\delta_P = 60$s while the threshold value $p_\text{thre} = 0.995$.
  
  To verify the superiority of our models in content caching, a comparison between the traditional caching method, LRU, and the prediction results based strategy is also carried out.
  We deploy \texttt{TGN-A} and its variants, e.g., \texttt{M2-STGN-A} and \texttt{M2-STGN-A+U}, as predictive models.
  Due to the trade-off between training speed and prediction performance for \texttt{M2-STGN-L+U} and \texttt{M2-STGN-L+U+SPE}, simulations based on them are executed as well.
  Moreover, testing in the inductive setting, we also conduct extensive ablation studies with \texttt{M2-STGN-A+U}, whose performance in preference prediction is the most superior, so as to examine the influence of different hyperparameters (i.e., different sizes of content supplement set and values of $\delta_P$ \& $p_{thre}$) on content caching.
  Notably, we adjust the size of the content supplement set by changing the duration before the starting time.

\subsection{Results Analysis}\label{5C}
  Table \ref{tab:my-table_1} demonstrates the prediction performance of our proposed TGN models as well as the baseline models. 
  It can be clearly observed that our models are able to yield better results in both transductive task and inductive task, and even the primitive models in \texttt{M1-STGN} outperform the counterparts of \texttt{TGN}. 
  Moreover, the superiority of \texttt{M1-STGN+U} and \texttt{M2-STGN+U} also proves the effectiveness of the \texttt{UsAttn} semantic aggregator. 
  In addition, due to the introduction of \texttt{SPE}, we can also find that the prediction capabilities of most models have been enhanced, especially for the variants of \texttt{TGN-L} and \texttt{TGN-M}. 
  Nevertheless, the results are of little avail when \texttt{SPE} collaborates with a complex model, e.g., \texttt{M2-STGN-A+U+SPE}.

 \begin{figure*}[tbp]
    \centering
      \subfloat[Different prediction threshold values.]{\includegraphics[width = 0.33\textwidth]{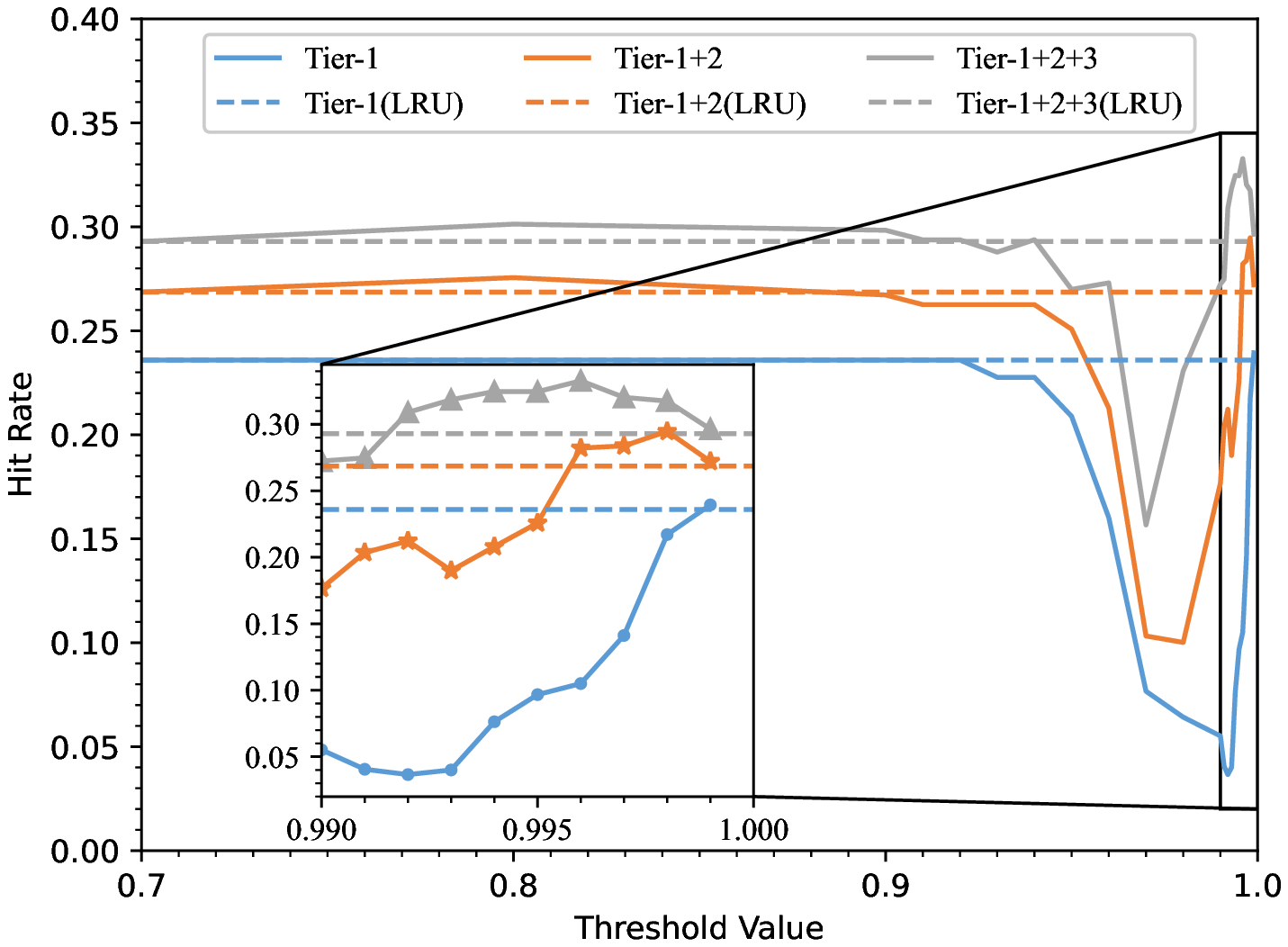}\label{fig:hypI_a}}
      \subfloat[Different popularity predicting periods.]{\includegraphics[width = 0.33\textwidth]{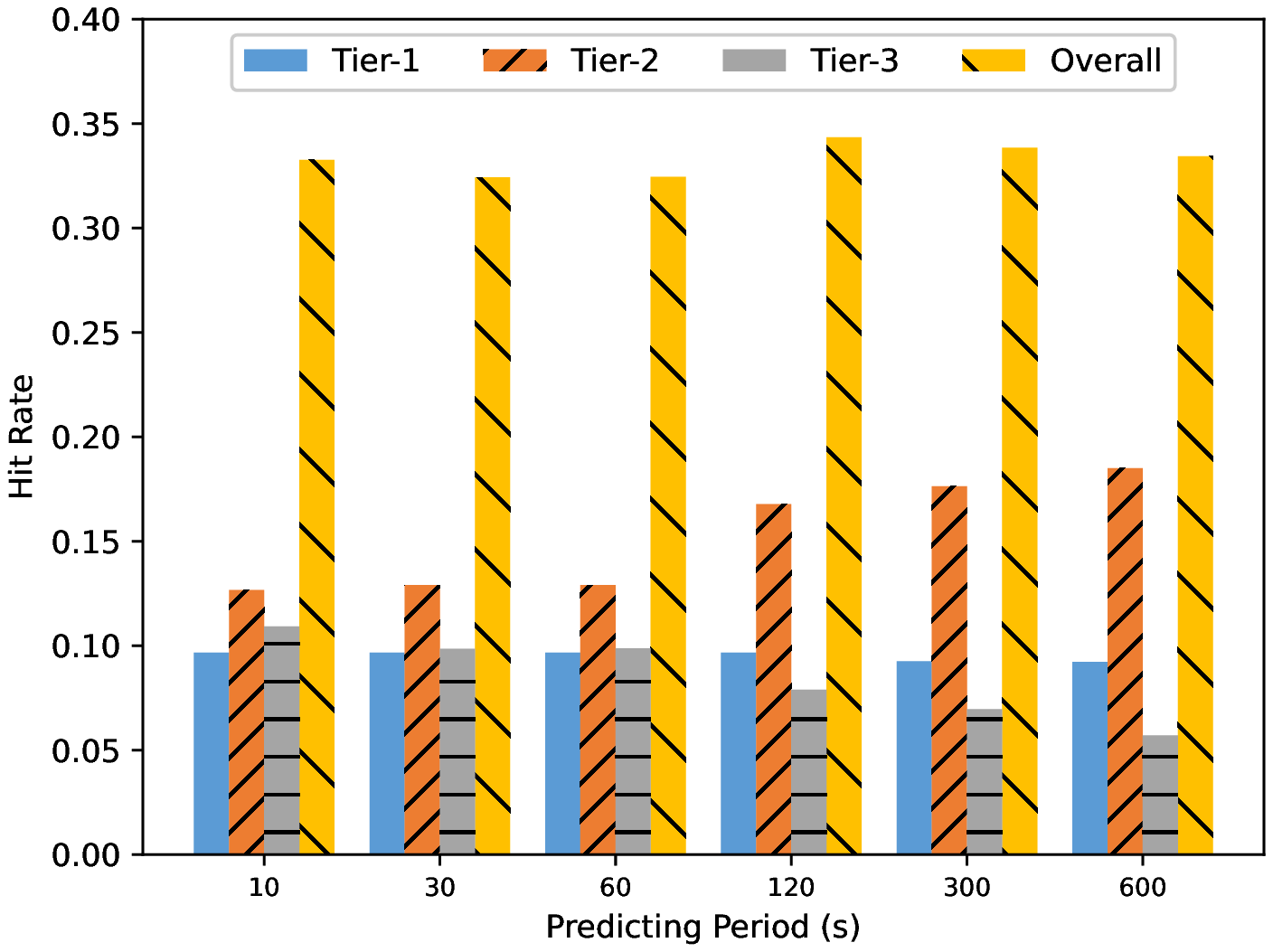}\label{fig:hypI_b}}
      \subfloat[Different numbers of candidate content.]{\includegraphics[width = 0.33\textwidth,height=0.25\textwidth]{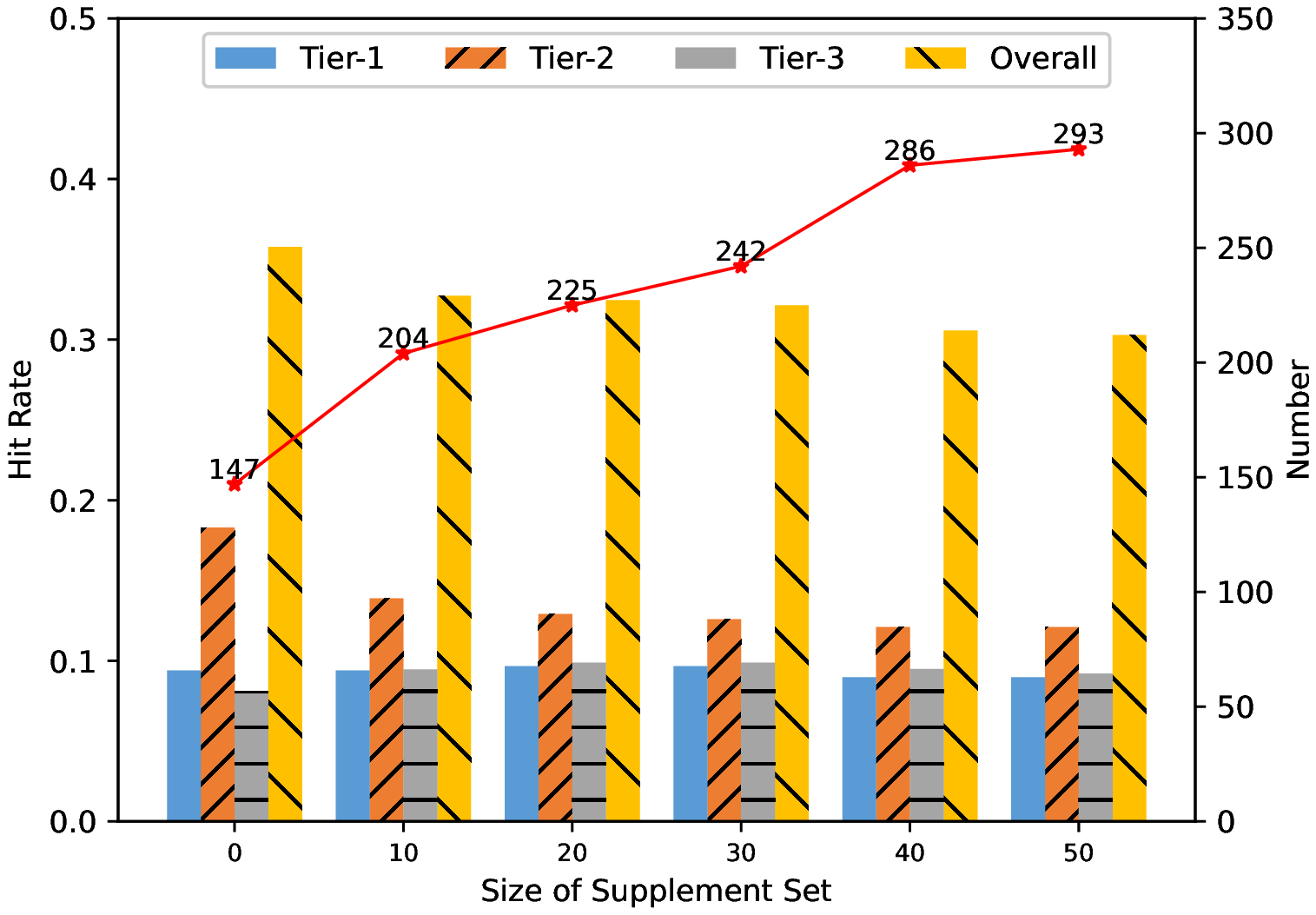}
      \label{fig:hypI_c}}
    \caption{The hit rate performance in the inductive setting with different hyperparameters for caching.}
    \vspace{-1.2em}
    \label{fig:hypI}
 \end{figure*}

  Fig. \ref{fig:layer} presents the prediction performance of \texttt{TGN-L} with different sizes of receptive field. 
  As the receptive field enlarges, the performance of \texttt{TGN-L} in both transductive and inductive tasks improves. 
  However, the average training time for obtaining a single model is gradually increasing as well.
  Compared with the corresponding results in Table \ref{tab:my-table_1} and Fig. \ref{fig:layer}, we can also discover that most variant models of \texttt{TGN-L} which try to have a deeper insight into the available information through either \texttt{UsAttn} or \texttt{SPE} are superior than methods that enlarge the receptive field (e.g., the stacking of GNN layers or an increase in the number of first-order neighbors) in both prediction performance and training speed.
  In order to achieve a comparable result to the \texttt{M2-STGN-L+U+SPE}, it takes \texttt{TGN-L} with 2 \texttt{TGAT} layers at least 4$\times$ more time.
  Obviously, our ``breadth-first approach'' for excavating the inherent relationships is more efficient. 

  Fig. \ref{fig:hitrate} compares the average hit rate of the caching strategy based on prediction results and LRU for the whole network architecture in both transductive and inductive tasks during 24 hours.
  It can be observed that relying on our proposed models, the overall caching performance of the prediction-based strategy is always better than LRU. 
  The improvement of prediction accuracy increases the cache hit rate as well. 
  In particular, the caching strategy based on \texttt{M2-STGN-A+U} surpasses other models, which is in line with the prediction performance.
  Moreover, considering the trade-off between training speed and the final performance in caching, caching with \texttt{M2-STGN-L+U+SPE} is also promising.
  Actually, we also conduct simulations for the other two baselines, i.e., \texttt{TGAT} and \texttt{DyRep}, but their poor performance results in too many false positive predictions, failing to distinguish the popularity of contents.

  Fig. \ref{fig:hypI} reveals the hit rate performance in the inductive setting with different hyperparameters for caching.
  It can be observed in Fig. \ref{fig:hypI}(a) that the caching performance with $p_\texttt{thre} < 0.8$ is equivalent to LRU.
  Since the caching strategy, introduced in Section \ref{cache}, decides the prioritization for contents with the same predicted popularity consistent with LRU, such an abnormal phenomenon implies that our model fails to distinguish the popularity of contents when adopting an improper threshold value.
  However, for a larger threshold, the caching gain from our model becomes more evident, especially when the threshold is close to 1, the strategy relying on our model outperforms the traditional LRU at all tiers.
  Surprisingly, a more frequent prediction operation does not always lead to an improvement in the hit rate. 
  The simulation results in Fig. \ref{fig:hypI}(b) present that when $\delta_p=120$s, our model brings the greatest gain to the cache task.
  On the other hand, Fig. \ref{fig:hypI}(c) shows that the size of candidate content also affects the final caching results.
  As the number of candidate content gradually increases, it gives rise to a declined overall hit rate, but is still superior to LRU.
  To sum up, these experiments demonstrate the robustness of the proposed methods.
    
\section{Conclusions}\label{sec7}
  In this paper, we have developed an STGN architecture to improve the performance of popularity prediction in a sparse dataset. 
  We have regarded the genres of the contents as semantic information and profiled users' intentions by attaching the semantics into the conventional TGN.
  The proposed STGN models have significantly ameliorated the user preference speculation performance.
  Furthermore, we have devised a \texttt{UsAttn} mechanism for a finer-grained semantic aggregation of diverse genres related to the same content.
  Meanwhile, an \texttt{SPE} function, targeting at assisting the association analysis in the attention-based graph learning, has been adopted as well.
  Due to the superior performance, the caching strategy based on our STGN model also wins a great improvement in cache hit rate under extensive simulations. 

  Finally, our model provides a paradigm for the fusion of semantics and AI models, where \texttt{UsAttn} suggests a novel method to aggregate multiple semantic information fine-grainedly and \texttt{SPE} answers the question about how to efficiently incorporate and utilize the aggregated semantic information with AI models.
  Meanwhile, besides the application in caching, we also believe that it has the potential to be generalized to other network architectures that desire AI models to integrate with semantic analysis, like intent-based network (IBN) \cite{9925251}.

\appendix[The Deduction of Semantic Positional Encoding]\label{APP:a}
    {
    As we discussed in Section \ref{SPE}, we regard the semantic features as a special kind of position in the semantic sphere.
    However, unlike the context sequence, handled by Transformer \cite{vaswani2017attention}, the semantic position is a multi-dimension representation.
    Compared to the cumbersome dimension-independent positional encoding method, we adopt a Fourier kernel-based method to solve this problem.

    In our GAT-alike \cite{velivckovic2017graph} mechanism, the inner-product self-attention belongs to one crucial element in calculating attention coefficients.
    Consider a continuous position mapping function $\Phi(\cdot)$, for two semantic positions $\textbf{\emph{x}}$ and $\textbf{\emph{y}}$, the inner-product between them can be formulated as $\langle \Phi(\textbf{\emph{x}}), \Phi(\textbf{\emph{y}})\rangle$.
    Notably, the relative value, which connotes the distance between two positions, is usually much more important than the absolute position value \cite{Xu2020Inductive}.
    Hence, we are more interested in learning a function that is related to the Euclidean distance between two points, i.e., $\sqrt{\| \textbf{\emph{y}} - \textbf{\emph{x}}\Vert^2 }$.
    Accordingly, we aim to define a function $\mathcal{K} (\cdot)$, where $\mathcal{K} (\textbf{\emph{x}}, \textbf{\emph{y}}) = \mathcal{K} (\textbf{\emph{y}}- \textbf{\emph{x}}) = \langle \Phi(\textbf{\emph{x}}), \Phi(\textbf{\emph{y}})\rangle$. It also implies that $\mathcal{K} (\cdot)$ is also a translation-invariant function.
    
    Next, we explain how to find an appropriate $\mathcal{K}(\cdot)$.
    Beforehand, we would like to present some fundamental lemmas on the Gram matrix, which are introduced in \cite{lanckriet2004learning},
    \begin{lemma}
      The Gram matrix is symmetric in the case where the real product is real-valued, and the corresponding entries are given by the inner product.
      \label{lemma_1}
    \end{lemma}
    \begin{lemma}
      The Gram matrix is positive-semidefinite.
      \label{lemma_2}
    \end{lemma}
    Therefore, by Lemma \eqref{lemma_1} and \eqref{lemma_2}, it can be observed that the result of $\mathcal{K} (\textbf{\emph{x}}, \textbf{\emph{y}})$ can be treated as an element of a Gram matrix and thus $\mathcal{K}$ is also positive semidefinite.
    Since any positive semidefinite function can be used as a kernel function \cite{lanckriet2004learning}, $\mathcal{K}(\cdot)$ can be taken as a kernel function as well.
    Moreover, the continuity of the mapping function $\Phi (\cdot)$ implies that the induced kernel $\mathcal{K}(\cdot)$ is continuous.
    Therefore, $\mathcal{K}$ shall satisfy the assumption of the Bochner's theorem \cite{rudin2017fourier}.
    \begin{lemma}
      \textit{(Bochner's theorem)}
        A continuous, translation-invariant 
        kernel $\mathcal{K}(a, b) = \mathcal{K}(a-b)$ on $\mathbb{R}^d$ is positive definite if
        and only if $\mathcal{K}(\cdot)$ is the Fourier transform of a non-negative measure.
    \end{lemma}

    Thus, Ref. \cite{rahimi2007random} asserts that $\mathcal{K}(\cdot)$ can be formulated as,
    \begin{equation}
        \mathcal{K}(a, b) = \mathcal{K}(a-b) = \int _{\mathbb{R}^d} \Pr(\omega)e^{j\omega (a-b)} d\omega = E_{\omega}[{\xi}(a) {\xi}(b)^{*}],
        \label{app_1}
    \end{equation}
    where ${\xi}(b)^{*}$ is the complex conjugate of ${\xi}(a)$. ${\xi}(a) = e^{j\omega a} = \cos ({\omega a}) + j \sin ({\omega a})$ and $\omega$ is drawn from $\Pr(\omega)$.
    Note that the kernel and probability distribution $\Pr(\omega)$ are real, so the imaginary part of ${\xi}(a) {\xi}(b)^{*}$ can be omitted, and \eqref{app_1} can be reformulated as
    \begin{equation}
        \begin{split}
            \mathcal{K}(a-b) &= E_{\omega}[\cos(\omega(a-b))] \\
            &= E_{\omega}[\cos(\omega a)\cos(\omega b)+\sin(\omega a)\sin(\omega b)]. 
        \end{split}
        \label{22}
    \end{equation}
    
    Therefore, by approximating expectation with the Monte Carlo integral \cite{rahimi2007random}, for each dimension $x_m$ and $y_m$ in $\textbf{\emph{x}}$ and $\textbf{\emph{y}}$, \eqref{22} implies that the inner-product of our kernel can be reformulated as
    \begin{equation}
      \begin{split}
        &\mathcal{K}_m(x_m-y_m) = \Phi({{x_m}})\cdot\Phi({{y_m}}) \\
      &\thickapprox \frac{1}{D_h}\sum_{d}\left(\cos(\omega_d {{x_m}})\cos(\omega_d {{y_m}})+\sin(\omega_d {{x_m}})\sin(\omega_d {{y_m}})\right) \\
      &= \frac{1}{D_h} \cos(\textbf{\emph{w}}_m (x_m -y_m)),
      \end{split}
    \end{equation}
    where $x_m$ and $y_m$ are the value of $m$-th dimension in $\textbf{\emph{x}}$ and $\textbf{\emph{y}}$, while $\omega_d$ is the $d$-th entry in $\textbf{\emph{w}}_m \in \mathcal{R}^{\frac{D_h}{2}}$.
    
    In other words, we can also denote the positional encoding function as $\Phi_m(x) = \frac{1}{\sqrt{D_h}}[\cos \textbf{\emph{w}}_m x_m || \sin \textbf{\emph{w}}_m x_m]^T $ for each dimension. 
    Then, we can encode the whole vector $\textbf{\emph{x}}$ with
    \begin{equation}
      \Phi(\textbf{\emph{x}}) = \frac{1}{\sqrt{D_h}}[\cos \textbf{\emph{W}}_{p}\textbf{\emph{x}} || \sin \textbf{\emph{W}}_{p}\textbf{\emph{x}}]^T.
    \end{equation}
    where $\textbf{\emph{W}}_{p} \in \mathcal{R}^{\frac{D_h}{2}\times D_m}$ is the stacking of $\textbf{\emph{w}}_m$, and ${D_m}$ is the dimension number of $\textbf{\emph{x}}$.

    Furthermore, we pay more attention to the Euclidean distance between the semantic positions, which suggests that $\mathcal{K}(\textbf{\emph{x}}, \textbf{\emph{y}})$ is the Fourier transform of a normal distribution with respect to $\omega$ \cite{rahimi2007random}.
    Thus, in our algorithm, we initialize entries of $\textbf{\emph{W}}_{p}$ in \eqref{pe_1} with a normal distribution while training models.
    }

  \bibliographystyle{IEEEtran}
  \bibliography{reference}

\begin{thebibliography}{10}
\providecommand{\url}[1]{#1}
\csname url@samestyle\endcsname
\providecommand{\newblock}{\relax}
\providecommand{\bibinfo}[2]{#2}
\providecommand{\BIBentrySTDinterwordspacing}{\spaceskip=0pt\relax}
\providecommand{\BIBentryALTinterwordstretchfactor}{4}
\providecommand{\BIBentryALTinterwordspacing}{\spaceskip=\fontdimen2\font plus
\BIBentryALTinterwordstretchfactor\fontdimen3\font minus
  \fontdimen4\font\relax}
\providecommand{\BIBforeignlanguage}[2]{{%
\expandafter\ifx\csname l@#1\endcsname\relax
\typeout{** WARNING: IEEEtran.bst: No hyphenation pattern has been}%
\typeout{** loaded for the language `#1'. Using the pattern for}%
\typeout{** the default language instead.}%
\else
\language=\csname l@#1\endcsname
\fi
#2}}
\providecommand{\BIBdecl}{\relax}
\BIBdecl

\bibitem{cisco2020cisco}
\BIBentryALTinterwordspacing
``Cisco annual internet report (2018--2023) white paper,'' Cisco, 2020.
  [Online]. Available:
  \url{https://www.cisco.com/c/en/us/solutions/collateral/executive-perspectives/annual-internet-report/white-paper-c11-741490.pdf}
\BIBentrySTDinterwordspacing

\bibitem{meybodi2022tedge}
Z.~Hajiakhondi~Meybodi, A.~Mohammadi, E.~Rahimian, S.~Heidarian, J.~Abouei, and
  K.~N. Plataniotis, ``{TEDGE-Caching}: Transformer-based edge caching towards
  {6G} networks,'' in \emph{Proc. ICC}, Seoul, South Korea, May 2022.

\bibitem{9134426}
Y.~Shi, K.~Yang, T.~Jiang, J.~Zhang, and K.~B. Letaief,
  ``Communication-efficient edge {AI}: {Algorithms} and systems,'' \emph{IEEE
  Commun. Surv. Tutor.}, vol.~22, no.~4, pp. 2167--2191, 2020.

\bibitem{ETSI}
\BIBentryALTinterwordspacing
``Mobile edge computing ({MEC}); {T}echnical requirements,'' ETSI, 2016.
  [Online]. Available:
  \url{https://www.etsi.org/deliver/etsi_gs/mec/001_099/002/01.01.01_60/gs_mec002v010101p.pdf}
\BIBentrySTDinterwordspacing

\bibitem{7194828}
C.~Yang, Y.~Yao, Z.~Chen, and B.~Xia, ``Analysis on cache-enabled wireless
  heterogeneous networks,'' \emph{IEEE Trans. Wirel. Commun.}, vol.~15, no.~1,
  pp. 131--145, 2016.

\bibitem{9427180}
Q.~Chen, W.~Wang, W.~Chen, F.~R. Yu, and Z.~Zhang, ``Cache-enabled multicast
  content pushing with structured deep learning,'' \emph{IEEE J. Sel. Areas
  Commun.}, vol.~39, no.~7, pp. 2135--2149, 2021.

\bibitem{9187344}
O.~Serhane, K.~Yahyaoui, B.~Nour, and H.~Moungla, ``A survey of {ICN} content
  naming and in-network caching in {5G} and beyond networks,'' \emph{IEEE
  Internet Things J.}, vol.~8, no.~6, pp. 4081--4104, 2021.

\bibitem{8172025}
W.~Liu, J.~Zhang, Z.~Liang, L.~Peng, and J.~Cai, ``Content popularity
  prediction and caching for {ICN}: {A} deep learning approach with {SDN},''
  \emph{IEEE Access}, vol.~6, pp. 5075--5089, 2018.

\bibitem{9234632}
Z.~Zhang and M.~Tao, ``Deep learning for wireless coded caching with unknown
  and time-variant content popularity,'' \emph{IEEE Trans. Wirel. Commun.},
  vol.~20, no.~2, pp. 1152--1163, 2021.

\bibitem{9846902}
T.~Zong, C.~Li, Y.~Lei, G.~Li, H.~Cao, and Y.~Liu, ``Cocktail edge caching:
  Ride dynamic trends of content popularity with ensemble learning,''
  \emph{{IEEE/ACM} Trans. Netw.}, vol.~31, no.~1, pp. 208--219, 2023.

\bibitem{zhou2020graph}
J.~Zhou, G.~Cui, S.~Hu, Z.~Zhang, C.~Yang, Z.~Liu, L.~Wang, C.~Li, and M.~Sun,
  ``Graph neural networks: A review of methods and applications,'' \emph{AI
  Open}, vol.~1, pp. 57--81, 2020.

\bibitem{wu2022graph}
S.~Wu, F.~Sun, W.~Zhang, X.~Xie, and B.~Cui, ``Graph neural networks in
  recommender systems: {A} survey,'' \emph{ACM Comput. Surv.}, vol.~55, no.~5,
  pp. 1--37, 2022.

\bibitem{skarding2021foundations}
J.~Skarding, B.~Gabrys, and K.~Musial, ``Foundations and modeling of dynamic
  networks using dynamic graph neural networks: {A} survey,'' \emph{IEEE
  Access}, vol.~9, pp. 79\,143--79\,168, May 2021.

\bibitem{9420319}
Y.~Fu, L.~Salaün, X.~Yang, W.~Wen, and T.~Q.~S. Quek, ``Caching efficiency
  maximization for device-to-device communication networks: {A} recommend to
  cache approach,'' \emph{IEEE Trans. Wirel. Commun.}, vol.~20, no.~10, pp.
  6580--6594, 2021.

\bibitem{zhu2022aoi}
J.~Zhu, R.~Li, G.~Ding, C.~Wang, J.~Wu, Z.~Zhao, and H.~Zhang, ``{AoI}-based
  temporal attention graph neural network for popularity prediction and content
  caching,'' \emph{IEEE Trans. on Cogn. Commun. Netw.}, 2022, {E}arly {A}ccess.

\bibitem{Xu2020Inductive}
D.~Xu, C.~Ruan, E.~Korpeoglu, S.~Kumar, and K.~Achan, ``Inductive
  representation learning on temporal graphs,'' in \emph{Proc. ICLR}, Virtual
  Edition, Apr./May 2020.

\bibitem{9220908}
J.~Liang, D.~Zhu, H.~Liu, H.~Ping, T.~Li, H.~Zhang, L.~Geng, and Y.~Liu,
  ``Multi-head attention based popularity prediction caching in social
  content-centric networking with mobile edge computing,'' \emph{IEEE Commun.
  Lett.}, vol.~25, no.~2, pp. 508--512, 2021.

\bibitem{liu2021contextualized}
Y.~Liu, S.~Yang, Y.~Xu, C.~Miao, M.~Wu, and J.~Zhang, ``Contextualized graph
  attention network for recommendation with item knowledge graph,'' \emph{IEEE
  Trans. Knowl. Data Eng.}, vol.~35, no.~1, pp. 181--195, 2023.

\bibitem{DBLP:conf/naacl/DevlinCLT19}
J.~Devlin, M.~Chang, K.~Lee, and K.~Toutanova, ``{BERT:} pre-training of deep
  bidirectional transformers for language understanding,'' in \emph{Proc.
  {NAACL-HLT}}, Minneapolis, Minnesota, USA, Jun. 2019.

\bibitem{pennington2014glove}
J.~Pennington, R.~Socher, and C.~D. Manning, ``Glove: Global vectors for word
  representation,'' in \emph{Proc. EMNLP}, Doha, Qatar, Oct. 2014.

\bibitem{vaswani2017attention}
A.~Vaswani, N.~Shazeer, N.~Parmar, J.~Uszkoreit, L.~Jones, A.~N. Gomez,
  {\L}.~Kaiser, and I.~Polosukhin, ``Attention is all you need,'' in
  \emph{Proc. NeurIPS}, Long Beach, CA, USA, Dec. 2017.

\bibitem{970573}
D.~Lee, J.~Choi, J.-H. Kim, S.~Noh, S.~L. Min, Y.~Cho, and C.~S. Kim, ``{LRFU}:
  {A} spectrum of policies that subsumes the least recently used and least
  frequently used policies,'' \emph{IEEE Trans. Comput.}, vol.~50, no.~12, pp.
  1352--1361, 2001.

\bibitem{trivedi2019dyrep}
R.~Trivedi, M.~Farajtabar, P.~Biswal, and H.~Zha, ``{DyRep}: {L}earning
  representations over dynamic graphs,'' in \emph{Proc. ICLR}, New Orleans, LA,
  USA, May 2019.

\bibitem{velivckovic2017graph}
P.~Veli{\v{c}}kovi{\'c}, G.~Cucurull, A.~Casanova, A.~Romero, P.~Lio, and
  Y.~Bengio, ``Graph attention networks,'' in \emph{Proc. ICLR}, Vancouver, BC,
  Canada, Apr./May 2018.

\bibitem{rossi2020temporal}
E.~Rossi, B.~Chamberlain, F.~Frasca, D.~Eynard, F.~Monti, and M.~Bronstein,
  ``Temporal graph networks for deep learning on dynamic graphs,'' in
  \emph{Proc. ICML 2020 Workshop on Graph Representation Learning}, Virtual
  Edition, Jul. 2020.

\bibitem{wang2018ripplenet}
H.~Wang, F.~Zhang, J.~Wang, M.~Zhao, W.~Li, X.~Xie, and M.~Guo, ``Ripplenet:
  {Propagating} user preferences on the knowledge graph for recommender
  systems,'' in \emph{Proc. CIKM}, Torino, Italy, Oct. 2018.

\bibitem{9409651}
Y.~Li, X.~Guo, W.~Lin, M.~Zhong, Q.~Li, Z.~Liu, W.~Zhong, and Z.~Zhu,
  ``Learning dynamic user interest sequence in knowledge graphs for
  click-through rate prediction,'' \emph{IEEE Trans. Knowl. Data Eng.},
  vol.~35, no.~1, pp. 647--657, 2023.

\bibitem{wang2019explainable}
X.~Wang, D.~Wang, C.~Xu, X.~He, Y.~Cao, and T.-S. Chua, ``Explainable reasoning
  over knowledge graphs for recommendation,'' in \emph{Proc. AAAI}, Honolulu,
  Hawaii, USA, Jan./Feb. 2019.

\bibitem{9354956}
H.~Mezni, D.~Benslimane, and L.~Bellatreche, ``Context-aware service
  recommendation based on knowledge graph embedding,'' \emph{IEEE Trans. Knowl.
  Data Eng.}, vol.~34, no.~11, pp. 5225--5238, 2022.

\bibitem{kitaev2020reformer}
N.~Kitaev, {\L}.~Kaiser, and A.~Levskaya, ``Reformer: The efficient
  transformer,'' in \emph{Proc. ICLR}, Virtual Edition, Apr./May 2020.

\bibitem{li2021learnable}
Y.~Li, S.~Si, G.~Li, C.-J. Hsieh, and S.~Bengio, ``Learnable fourier features
  for multi-dimensional spatial positional encoding,'' in \emph{Proc. NeurIPS},
  Virtual Edition, Dec. 2021.

\bibitem{46840}
N.~J. Parmar, A.~Vaswani, J.~Uszkoreit, L.~Kaiser, N.~Shazeer, A.~Ku, and
  D.~Tran, ``Image transformer,'' in \emph{Proc. ICML}, Stockholm, Sweden, Jul.
  2018.

\bibitem{ayoub2018energy}
O.~Ayoub, F.~Musumeci, M.~Tornatore, and A.~Pattavina, ``Energy-efficient
  video-on-demand content caching and distribution in metro area networks,''
  \emph{IEEE Trans. Green Commun. Netw.}, vol.~3, no.~1, pp. 159--169, Mar.
  2019.

\bibitem{8425746}
B.~Chen and C.~Yang, ``Caching policy for cache-enabled d2d communications by
  learning user preference,'' \emph{IEEE Trans. Commun.}, vol.~66, no.~12, pp.
  6586--6601, 2018.

\bibitem{chung2014empirical}
J.~Chung, C.~Gulcehre, K.~Cho, and Y.~Bengio, ``Empirical evaluation of gated
  recurrent neural networks on sequence modeling,'' in \emph{Proc. NeurIPS},
  Montreal, QC, Canada, Dec 2014.

\bibitem{rahimi2007random}
A.~Rahimi and B.~Recht, ``Random features for large-scale kernel machines,'' in
  \emph{Proc. NeurIPS}, Vancouver, BC, Canada, Dec. 2007.

\bibitem{jawahar2019does}
G.~Jawahar, B.~Sagot, and D.~Seddah, ``What does {BERT} learn about the
  structure of language?'' in \emph{Proc. ACL}, Florence, Italy, Jul./Aug.
  2019.

\bibitem{9925251}
A.~Leivadeas and M.~Falkner, ``A survey on {Intent-Based Networking},''
  \emph{IEEE Commun. Surveys Tuts.}, vol.~25, no.~1, pp. 625--655, 2023.

\bibitem{lanckriet2004learning}
G.~R. Lanckriet, N.~Cristianini, P.~Bartlett, L.~E. Ghaoui, and M.~I. Jordan,
  ``Learning the kernel matrix with {semidefinite} programming,'' \emph{J.
  Mach. Learn Res.}, vol.~5, no. Jan, pp. 27--72, 2004.

\bibitem{rudin2017fourier}
W.~Rudin, \emph{Fourier analysis on groups}.\hskip 1em plus 0.5em minus
  0.4em\relax Courier Dover Publications, 2017.

\end{thebibliography}
\end{document}